\documentclass{sbc2023}

\usepackage{graphicx}
\usepackage[misc,geometry]{ifsym} 
\usepackage{fontspec}
\usepackage{float}
\usepackage{color}
\usepackage{xcolor}
\usepackage{hyperref} 
\usepackage{aas_macros}
\usepackage[bottom]{footmisc}
\usepackage{supertabular}
\usepackage{subcaption}
\usepackage{afterpage}
\usepackage{url}
\usepackage{pifont}
\usepackage{multicol}
\usepackage{multirow}
\usepackage{longtable}
\usepackage{eso-pic}

\setcitestyle{square}
\definecolor{lightgray}{rgb}{0.92, 0.92, 0.92}
\definecolor{unilogo}{rgb}{0.16, 0.26, 0.58}
\definecolor{maillogo}{rgb}{0.58, 0.16, 0.26}
\definecolor{darkblue}{rgb}{0.0,0.0,0.0}
\hypersetup{colorlinks,breaklinks,
            linkcolor=darkblue,urlcolor=darkblue,
            anchorcolor=darkblue,citecolor=darkblue}

\providecommand{\faEnvelopeO}{*}
\providecommand{\faCreativeCommons}{}

\jid{JBCS}
\jtitle{Journal of the Brazilian Computer Society, 202X, XX:1, }
\doi{10.5753/jbcs.202X.XXXXXX}
\copyrightstatement{This work is licensed under a Creative Commons Attribution 4.0 International License}
\jyear{202X}

\title[Aggregating Neighbor Embedding Projection and Rank-Based Manifold Learning for Image Retrieval]{Aggregating Neighbor Embedding Projection and Rank-Based Manifold Learning for Image Retrieval}

\author[Kawai et al. 2026]{
\affil{\textbf{Vinicius Atsushi Sato Kawai}~\textcolor{blue}{\faEnvelopeO}~~[~\textbf{S\~{a}o Paulo State University (UNESP), Rio Claro, Brazil}~|\href{mailto:vinicius.kawai@unesp.br}{~\textbf{\textit{vinicius.kawai@unesp.br}}}~]}

\affil{\textbf{Gustavo Rosseto Leticio}~\textcolor{blue}{\faEnvelopeO}~~[~\textbf{S\~{a}o Paulo State University (UNESP), Rio Claro, Brazil}~|\href{mailto:gustavo.leticio@unesp.br}{~\textbf{\textit{gustavo.leticio@unesp.br}}}~]}

\affil{\textbf{Lucas Pascotti Valem}~~\textcolor{orange}{\faEnvelopeO}~[~\textbf{University of S\~{a}o Paulo (USP), S\~{a}o Carlos, Brazil}~|\href{mailto:lucas@icmc.usp.br}{~\textbf{\textit{lucas@icmc.usp.br}}~]}}

\affil{\textbf{Daniel Carlos Guimarães Pedronette}~\textcolor{blue}{\faEnvelopeO}~~[~\textbf{S\~{a}o Paulo State University (UNESP), Rio Claro, Brazil~}|\href{mailto:daniel.pedronette@unesp.br}{~\textbf{\textit{daniel.pedronette@unesp.br}}}~]}

}

\begin{document}

\AddToShipoutPictureFG*{
  \AtPageUpperLeft{
    \raisebox{-0.5cm}{
      \makebox[\paperwidth][c]{
        \fcolorbox{black}{gray!12}{
          \parbox{0.85\textwidth}{\centering\small\bfseries\itshape
          Preprint of paper accepted for publication in the Journal of the Brazilian Computer Society (JBCS).}
        }
      }
    }
  }
}

\begin{frontmatter}
\maketitle

\begin{mail}
Department of Statistics, Applied Mathematics, and Computing (DEMAC), S\~{a}o Paulo State University (UNESP), Av. 24 A, Rio Claro, SP, 13506-900, Brazil.
\end{mail}

\begin{mail2}
Institute of Mathematics and Computer Science (ICMC), University of S\~{a}o Paulo (USP), Av. Trabalhador São-carlense, 400 - Centro, São Carlos, SP, 13566-590, Brazil.
\end{mail2}

\begin{dates}
\small{\textbf{Received:} DD Month YYYY~~~$\bullet$~~~\textbf{Accepted:} DD Month YYYY~~~$\bullet$~~~\textbf{Published:} DD Month YYYY}

\end{dates}

\begin{abstract}
\textbf{Abstract.~}
Content-based image retrieval (CBIR) has evolved significantly with the advent of deep learning models, yet effectively ranking similar images remains a challenging task, particularly in high-dimensional feature spaces where pairwise distance measures often fail to capture complex contextual relationships and the semantic gap between visual features and high-level concepts persists. In this scenario, manifold learning and rank-based refinement methods have emerged as complementary strategies, respectively improving feature representations and exploiting contextual information embedded in ranked lists, such as neighborhood relationships among images. However, combining projection-based and rank-based strategies to exploit their complementary properties remains a challenging research problem.
To address this, this paper proposes a framework that combines neighbor embedding projections with rank-based manifold learning through rank aggregation. Specifically, Uniform Manifold Approximation and Projection (UMAP) is used to generate alternative low-dimensional feature representations, while ranked lists obtained from UMAP projections and rank-based re-ranking methods are combined using the Borda Count aggregation strategy. The experimental evaluation was conducted on several public datasets using deep learning features extracted from ResNet152, Swin Transformer, and DINOv2 models.
The results show that the proposed approach can improve retrieval effectiveness in several scenarios, particularly when the baseline representation struggles to achieve high precision values. In addition, the aggregation strategy often improves the quality of the top-ranked positions, leading to competitive Mean Average Precision (MAP) and Precision values across different datasets and feature extractors. These findings suggest that combining projection-based and rank-based manifold learning strategies through rank aggregation can provide complementary contextual information for image retrieval tasks.
\end{abstract}

\begin{keywords}
Image Retrieval, Manifold Learning, Rank Aggregation, Projection, Re-ranking
\end{keywords}


\end{frontmatter}


\section{Introduction}
\label{sec:intro}

Content-Based Image Retrieval (CBIR) remains a fundamental challenge in computer vision due to the inherent difficulty in translating raw pixel data into high-level human perception~\citep{long2003fundamentals}.
These systems address the task of retrieving images from a database based on their visual content, such as color, texture, shape, or deep features, rather than relying only on textual metadata. They aim to return the most visually similar or semantically relevant images to a query image, supporting diverse applications such as medical imaging, remote sensing, surveillance, and multimedia search~\citep{muller2004review, vharkate2022fusion}.

While traditional CBIR approaches typically rely on pairwise distance measures, such as Euclidean distance~\citep{PaperExNNAlg_PKDD2007}, these metrics often fail to capture the complex contextual and geometric relationships between images in high-dimensional feature spaces~\citep{valem2022novel}. A primary obstacle is the well-known semantic gap~\citep{barz2021content}, where low-level visual features, such as color and texture, do not necessarily coincide with the abstract concepts or categories a user intends to retrieve. Furthermore, the \textit{curse of dimensionality}~\citep{kouiroukidis2011effects} in modern feature vectors introduces significant computational costs and memory requirements, which can lead to a deterioration in retrieval quality if the underlying manifold structure of the data is not properly preserved.

In this sense, recent advancements in the field have shifted the focus toward leveraging deep learning architectures, such as ResNet152, Swin Transformer, and DINOv2~\citep{oquab2023dinov2, paperRESNET}, to extract more discriminative and effective visual features. Despite these improvements in feature representation, effectively ranking images remains a modern bottleneck, particularly in fine-grained retrieval scenarios where inter-class similarity is high and intra-class variability is significant.
This challenge becomes even more pronounced in unsupervised settings, where no labeled data is available to guide feature selection or parameter tuning. In such cases, retrieval methods rely solely on the intrinsic data distribution and underlying geometric structure, making the design of robust and adaptive similarity strategies considerably more difficult~\citep{chen2003unsupervised, pereira2024unsupervised}.

In this scenario, manifold learning and rank-based refinement techniques have emerged as effective strategies to improve retrieval performance. Projection-based methods, such as Uniform Manifold Approximation and Projection (UMAP)~\citep{mcinnes2018umap}, aim not only to reduce dimensionality but also to generate more discriminative embeddings by preserving intrinsic geometric relationships~\citep{kawai2025semi}. In parallel, re-ranking, often associated with unsupervised affinity learning, became an effective post-processing strategy to refine initial ranked lists or affinity matrices that encode pairwise similarity relationships between data objects~\citep{valem2018unsupervised, bai2019re}.
Existing approaches for context-aware similarity learning can be broadly categorized into diffusion processes, rank-based strategies, and deep learning-based methods~\citep{pereira2024unsupervised}. To improve these affinities, such techniques leverage the underlying data manifold, i.e., the intrinsic geometric structure of high-dimensional data, enabling the incorporation of global contextual information beyond simple pairwise comparisons. As a result, they produce more consistent similarity relationships and improved retrieval accuracy. By capturing different contextual relationships, re-ranking strategies provide complementary views of the data that can be further exploited to enhance retrieval effectiveness. These relationships are typically modeled through mathematical constructs such as weighted affinity graphs, transition matrices, and reciprocal k-nearest neighbor (kNN) graphs, which explicitly represent the dataset manifold~\citep{pereira2024unsupervised}.

While both projection-based and rank-based strategies have demonstrated effectiveness, the way their resulting rankings can be combined to exploit complementary information remains relatively underexplored. Conversely, the literature has extensively investigated other forms of complementarity, particularly at the feature level, where multiple descriptors or deep models are combined to enhance representation quality~\citep{zhou2015augmented, vharkate2022fusion, belalia2025enhanced}. However, comparatively less attention has been given to the fusion of ranked lists generated after manifold-based transformations, where complementary information emerges from distinct ranking processes rather than from the feature space itself.

Moreover, existing approaches based on graph modeling, pair-wise, list-wise and iterative re-ranking have demonstrated the potential of leveraging contextual relationships within ranked lists~\citep{valem2023rank, xiao2025locore}. Despite these advances, integrating heterogeneous sources of information, such as projection-based representations and rank-based refinements, into a unified and effective framework remains a challenging problem, especially when aiming to preserve the quality of top-ranked results while reducing redundancy across rankings.

Motivated by these challenges, this paper proposes a unified framework that integrates projection-based and rank-based manifold learning through a rank aggregation strategy. Specifically, we combine ranked lists obtained from UMAP-based projections and rankings refined from the original feature space using re-ranking methods. These independent ranking processes are aggregated using the Borda Count~\citep{emerson2013original} strategy, followed by an optional post re-ranking step for additional refinement.
This work extends our previous study~\citep{kawai2024} by moving from a sequential combination of projection and re-ranking to a formulation based on independent ranking generation and aggregation. This design enables a more effective exploitation of complementary information between different ranking processes.

The central hypothesis of this work is that rank aggregation can capture complementary contextual information between projection-based and rank-based manifold learning techniques. By combining geometric relationships from projection methods with contextual refinements from ranked lists, the proposed approach aims to improve retrieval effectiveness, particularly in scenarios where the baseline representation is not sufficiently discriminative.

In this manner, the proposed framework is evaluated across multiple datasets and feature extraction models, demonstrating that the aggregation of manifold-based rankings can provide competitive performance and, in several cases, improve the quality of top-ranked retrieval results.

The remainder of this paper is organized as follows: Section~\ref{sec:rel_work} discusses related work, Section~\ref{sec:proposed_approach} presents the proposed approach and its components, Section~\ref{sec:experimental_eval} details the experimental evaluation, and Section~\ref{sec:conclusion} concludes the paper with insights and future directions.

\section{Related Work}
\label{sec:rel_work}

Recent advances in machine learning and information retrieval have led to the development of various strategies for improving ranking quality in a large variety of retrieval scenarios. While traditional similarity search methods rely on feature-space distance computations, alternative approaches have explored structural properties of data manifolds, contextual relationships in ranked lists, and rank aggregation techniques to enhance retrieval effectiveness.

In particular, manifold learning techniques enable more meaningful feature space transformations by preserving the geometric structure of high-dimensional data. These methods have been widely used not only in retrieval tasks but also in data visualization~\citep{wang2021understanding, rafieian2023improving}, biological analysis~\citep{Becht2019, dorrity2020dimensionality, trozzi2021umap}, and unsupervised learning applications~\citep{pedronette2019multimedia}. Meanwhile, rank-based refinement methods~\citep{valem2023rank, sabahi2024refinerhash} improve retrieval quality by leveraging ranking structures rather than direct feature comparisons, refining initial rankings based on contextual similarities. Finally, rank aggregation approaches~\citep{pedronette2012combining, sculley2007rank} fuse rankings from multiple retrieval strategies, improving the overall retrieval output by integrating complementary ranking information. 

In this sense, this section provides an overview of these three key directions in retrieval optimization. Section~\ref{subsec:rel_work_neighbor_embedding_projection} discusses manifold learning for neighbor embedding projection, emphasizing its role in dimensionality reduction and structured feature transformation. Section~\ref{subsec:rel_work_reranking} explores rank-based manifold learning and re-ranking techniques, highlighting how ranking structures can refine retrieval effectiveness. Finally, Section~\ref{subsec:rel_work_rank_agg} reviews rank aggregation methods, discussing different fusion strategies.


\subsection{Manifold Learning for Neighbor Embedding Projection}
\label{subsec:rel_work_neighbor_embedding_projection}

Manifold learning techniques seek to represent high-dimensional data in a lower-dimensional space while preserving essential geometric properties. These methods are particularly useful for tasks where data is assumed to lie on a complex, non-Euclidean structure, making direct similarity comparisons in the original space less meaningful.

Several methods have been developed to address this problem. Isomap~\citep{tenenbaum2000global} constructs a neighborhood graph and estimates geodesic distances to capture global data structure, making it effective for preserving long-range relationships. Locally Linear Embedding (LLE)~\citep{Roweis2000} reconstructs local neighborhoods using linear coefficients, ensuring smooth transitions between neighboring points. Laplacian Eigenmaps~\citep{belkin2001laplacian} optimize embeddings by minimizing distortion in local connectivity, leveraging spectral graph theory to map data into a reduced space while maintaining adjacency relations.

Probabilistic approaches such as Stochastic Neighbor Embedding (SNE)~\citep{hinton2002stochastic} and its refined version, t-SNE~\citep{PapertSNE_vanDerMaaten2008}, model similarities as probability distributions, emphasizing local neighborhood preservation. These methods excel at clustering structures but can struggle with global organization. Diffusion Maps~\citep{coifman2006diffusion} employ a Markov process on the data graph to extract meaningful diffusion distances, capturing intrinsic relationships in data that are robust to noise.

A more recent and widely adopted approach is Uniform Manifold Approximation and Projection (UMAP)~\citep{mcinnes2018umap}, which builds a topological representation of data by constructing a fuzzy simplicial complex. Its computational efficiency and enhanced neighborhood consistency have led to successful applications in various domains, including genomic analysis~\citep{diaz2019umap}, clustering tasks~\citep{allaoui2020considerably, sanchez2023combination}, classification~\citep{milovsevic2022application} and more recently content-based image retrieval~\citep{leticio2024manifold, kawai2024}.

Each of these methods presents unique advantages depending on the data distribution and task requirements~\citep{8851280}. The choice of projection technique significantly influences retrieval effectiveness by shaping how similarity is interpreted in the reduced space, ultimately impacting the quality of ranking-based search strategies.


\subsection{Rank-Based Manifold Learning and Re-Ranking}
\label{subsec:rel_work_reranking}

Rank-based manifold learning techniques aim to enhance retrieval effectiveness by leveraging the contextual relationships embedded in ranked lists. Unlike traditional feature-space methods, rank-based approaches focus on refining the ordering of retrieved items, ensuring better consistency and accuracy in similarity assessments.

One of the primary strategies in rank-based learning involves the construction of graphs or hypergraphs to model the relationships among ranked lists. Reciprocal k-Nearest Neighbors (rkNN) graphs utilize mutual neighborhood information to establish more reliable similarity connections, thus reducing noise from ambiguous rankings. These graphs leverage robust similarity information by emphasizing reciprocal relations, which are less sensitive to outliers~\citep{pedronette2018unsupervised, pedronette2014unsupervised}.

Hypergraph-based ranking extends this notion by representing higher-order relationships among multiple ranked lists. In this way, instead of pairwise connections, hypergraphs capture group-wise similarities, allowing the propagation of ranking information across interconnected nodes. Some hypergraph-based approaches include HyperSSR~\citep{jing2018hyperssr}, LHRR~\citep{pedronette2019multimedia} and RFE~\citep{valem2023rank}.

Another interesting direction is the use of diffusion processes. These methods propagate similarity information through the graph structure, gradually refining the ranked lists. Techniques like Locally Constrained Diffusion Process (LCDP)~\citep{yang2009locally} ensure that the diffusion is restricted within local neighborhoods, preserving locality while enhancing rankings consistency. Similarly, several diffusion-based methods have been proposed~\citep{jiang2011unsupervised, 6619018}.

\subsection{Rank Aggregation for Retrieval Improvement}
\label{subsec:rel_work_rank_agg}

Rank aggregation techniques aim to combine multiple ranked lists into a single consensus ranking. This approach leverages the complementary information present in different ranking sources, leading to more accurate and robust retrieval results. In content-based image retrieval (CBIR), rank aggregation is particularly useful for integrating rankings obtained from diverse feature representations or retrieval models, enhancing the overall retrieval effectiveness.

In this manner, several methods have been proposed to achieve this aggregation. Borda Count~\citep{emerson2013original} is a classic approach that assigns scores to each item based on its position across the ranked lists, summing these scores to determine the final order. Markov Chain-based rank fusion~\citep{liu2007supervised} models the ranking process as a Markov chain, where transition probabilities reflect the relative importance of each item, resulting in a stationary distribution that represents the consensus ranking.

In addition, Agglomerative Rank Aggregation~\citep{pedronette2012combining} has been proposed as a hierarchical approach that combines multiple rank aggregation methods. By cascading the outputs from one layer as inputs to the next, this approach captures complex relationships among rankings and improves the robustness of the final ordering, leveraging the strengths of multiple aggregation techniques~\citep{pedronette2011exploiting, pedronette2013image}.

From this perspective, manifold learning techniques play a crucial role in enhancing feature representation through neighbor embedding projections, while rank-based methods refine retrieval by leveraging contextual information. Rank aggregation further improves retrieval consistency by consolidating multiple ranking perspectives. Building on these advances, this paper proposes a unified approach that integrates neighbor embedding projections with rank-based re-ranking through an effective rank aggregation strategy, aiming to enhance retrieval accuracy.


\section{Proposed Approach}
\label{sec:proposed_approach}

This section presents our proposed approach, and it is divided into the following sections: 
Section~\ref{subsec:overview} presents an overview, describing each step of our method; 
Section~\ref{subsec:formal_definitions} introduces the formal definitions and notation used;
Section~\ref{subsec:neighbor_embedding_projection} focuses on the use of UMAP for neighbor embedding projection;
Section~\ref{subsec:rank_based_manifold} discusses the rank-based manifold learning methods; Finally, Section~\ref{subsec:rank_agg} details the rank aggregation step.

\begin{figure*}[ht!]
    \centering
    \includegraphics[width=.99\textwidth]{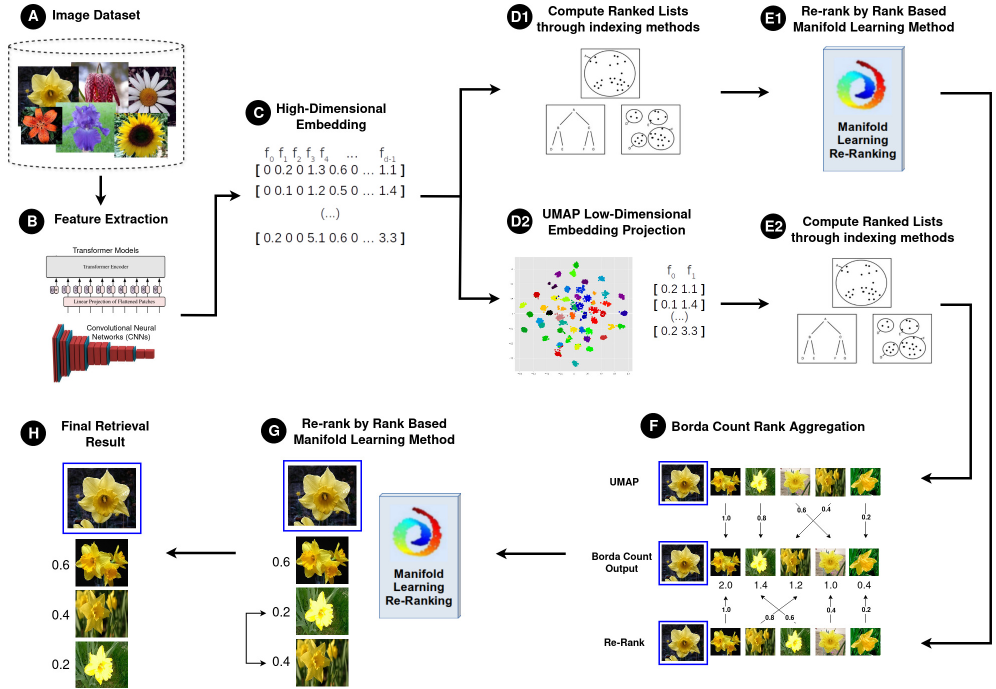}
    \vspace{5mm}
    \caption{Proposed Approach: Combined Manifold by Ranking and Projection through Rank Aggregation for Content Based Image Retrieval Tasks.}
    \label{fig:metodo}
\end{figure*}

\subsection{Overview}
\label{subsec:overview}

The proposed approach combines rank-based manifold learning with a neighbor embedding projection technique to enhance content-based image retrieval. The methodology follows a structured sequence of steps, as illustrated in Figure \ref{fig:metodo} and described below:

\noindent $\bullet$ \textbf{Image Dataset and Feature Extraction [A-C]}: The process begins with an image dataset from which deep learning models, such as CNNs~\citep{paperRESNET} and Transformers~\citep{paperSWIN-TF, oquab2023dinov2}, extract high-dimensional feature representations. These features serve as the foundation for subsequent retrieval steps.

\noindent $\bullet$ \textbf{Initial Ranking and Dimensionality Reduction [D1-D2]}: The extracted high-dimensional features are first used to generate an initial ranked list based on Euclidean distances using an indexing method (e.g., BallTree). In parallel, Uniform Manifold Approximation and Projection (UMAP) is applied to project these features into a lower-dimensional space while preserving both local and global structures. This step results in more compact and improved representations.

\noindent $\bullet$ \textbf{Re-Ranking and Secondary Ranking [E1-E2]}: The initial ranked lists are refined using rank-based manifold learning methods, enhancing retrieval quality by prioritizing more relevant images. Simultaneously, a new ranked list is generated using UMAP's reduced features with the same indexing method, providing an alternative similarity ranking.

\noindent $\bullet$ \textbf{Aggregation of Ranked Lists [F]}: The ranked lists obtained from the original feature space and the UMAP-reduced space are combined using an aggregation strategy based on the Borda Count method. This process merges ranking information from both representations into a single refined list.

\noindent $\bullet$ \textbf{Post Re-Rank [G]}: A re-ranking step is applied to the aggregated ranks, further refining the retrieval results and improving the prioritization of relevant images.

\noindent $\bullet$ \textbf{Final Retrieval Results [H]}: The optimized ranked lists are used to retrieve the most similar images for each query, leveraging the combined benefits of dimensionality reduction, ranking fusion, and manifold-based re-ranking.


\subsection{Formal Definitions}
\label{subsec:formal_definitions}

Mostly based on our initial work~\citep{kawai2024}, this section presents the formal definitions used for each task of our proposed approach.

\noindent $\bullet$ \textbf{Image Retrieval Task} 

The goal of the image retrieval task is to identify and return images from a collection $\mathcal{C}$ that are most similar to a given query image $x_q$. Retrieval is typically performed based on feature representations extracted from images, which encode their content in a high-dimensional space. Formally, a feature descriptor is defined as a function $f:\mathcal{C} \rightarrow \mathbb{R}^d$ that maps an image to a $d$-dimensional feature vector, such that $\mathbf{x_i} = f(x_i)$ and $\mathbf{x_i} = [x_{i1}, x_{i2}, \dots, x_{id}]$, where $x_{ij} \in \mathbb{R}$ represents the $j$-th feature of image $x_i$.  

Thus, the image collection can be expressed as $\mathcal{C} = \{x_1, x_2, \dots, x_n\}$, where each image $x$ is represented by a corresponding feature vector $\mathbf{x}$. The set $\mathcal{X} \subset \mathbb{R}^d$ consists of $n$ points in a $d$-dimensional Euclidean space $\mathbb{R}^d$, ensuring that each image representation $\mathbf{x_i} \in \mathcal{X}$.  

The similarity of the two images is determined by measuring the distance between their feature vectors. A distance function $\rho:\mathbb{R}^d \times \mathbb{R}^d \rightarrow \mathbb{R}^+$, commonly the Euclidean distance, is used to quantify this similarity. Therefore, the distance between images $x_i$ and $x_j$ is given by $\rho(\mathbf{x_i}, \mathbf{x_j})$, where $\mathbf{x} \in \mathbb{R}^d$ represents the feature vector of an image $x$.

The nearest neighbor search problem consists of identifying the element $\mathcal{N}({\mathbf{x_q}})$ within the set $\mathcal{X}$ that has the smallest distance $\rho$ to the query image $\mathbf{x_q}$. This concept can be extended to the $k$ nearest neighbors $\mathcal{N}({\mathbf{x_q}}, k)$, which includes the $k$ closest elements to the query. Consequently, the ranked list $\tau_q$ is defined as a permutation of $(x_1, x_2, \dots, x_n)$ within $\mathcal{N}({\mathbf{x_q}}, k)$, where $\tau_q (x_i)$ represents the position (or rank) of image $x_i$ in the ranked list $\tau_q$. If $x_i$ appears before $x_j$ in the ranking for $x_q$, i.e., $\tau_q (x_i) < \tau_q (x_j)$, then it follows that $\rho(\mathbf{x_{q}, x_{i}}) \leq \rho(\mathbf{x_{q}, x_{j}})$.

\noindent $\bullet$ \textbf{Rank-based Manifold Learning}

Rank-based manifold learning techniques leverage ranked lists $\mathcal{T} = [\tau_1, \tau_2, \dots, \tau_n]$ to reveal meaningful patterns within the data. These methods operate in an unsupervised manner, refining the initial rankings to obtain a more representative set of ranked lists, denoted as $\mathcal{T}_m$. This refinement process can be formally described by the function $f_m$:  
\begin{equation}
    \mathcal{T}_m = f_m (\mathcal{T}) 
\end{equation}
Since these methods refine the ranked lists, they are commonly categorized as re-ranking approaches.

\noindent $\bullet$ \textbf{Neighbor Embedding Projection}

Similarity estimation between images using pairwise distance metrics, such as Euclidean distance in the original high-dimensional feature space, often fails to preserve the underlying manifold structure of the data. Additionally, high-dimensional spaces introduce significant challenges, including increased computational cost, higher memory requirements, and potential deterioration in retrieval quality.

To overcome these limitations, we adopt a projection method based on the neighbor embedding projection framework, which leverages spatial relationships in a lower-dimensional space. Formally, a projection function is defined as $P:\mathbb{R}^d \rightarrow \mathbb{R}^q$, where $q \ll d$, typically with $q = 2$. This function maps a set $\mathcal{X}$ from the high-dimensional space to a lower-dimensional representation $\mathcal{X}'$, such that $\mathcal{X}' = P(\mathcal{X})$. Following this notation, we define an image representation from the low-dimensional space as $\mathbf{x_i'} \in \mathcal{X}'$ and the distance between the low-dimensional representations as $\rho'$. In this sense, $\tau_{q}'$ denotes the permutation of $(x_1, x_2, \dots, x_n)$ within $\mathcal{N}(\mathbf{x}_q', k)$. Similarly to the image retrieval notation, if $x_i$ appears before $x_j$ in the ranking for $x_q$, i.e., $\tau_q'(x_i) < \tau_q'(x_j)$, then $\rho'(\mathbf{x}_q', \mathbf{x}_i') < \rho'(\mathbf{x}_q', \mathbf{x}_j')$.

\noindent $\bullet$ \textbf{Rank Aggregation}

In the context of our proposed approach, the rank aggregation task combines rankings generated from different sources to produce a consolidated rank that better reflects the similarity between images. Formally, for a query image $x_q$, we denote:

\begin{itemize}
    \item $\tau_q$ as the ranking generated by ordering the distances $\rho$ between the feature vector $\mathbf{x}_q$ and the feature vectors of other images in the dataset ($\mathbf{x}_i \in \mathcal{X}$). 
    \item $\tau_q'$ as the ranking generated by ordering the distances $\rho'$ between the two-dimensional feature vectors obtained after the neighbor embedding projection.
    \item $\tau_{m_q}$ as the ranking obtained after the refinement process from rank-based manifold learning approaches.
    \item $\tau_q''$ as the aggregated ranking that combines the information from both $\tau_{m_q}$ and $\tau_q'$.
\end{itemize}

The rank aggregation function is then defined as:

\begin{equation}
    \mathcal{T}'' = g(\mathcal{T'}, \mathcal{T}_m)
\end{equation}

\noindent where $g$ is a function that consolidates the rankings by considering the relative positions of images in both $\mathcal{T}'$ and $\mathcal{T}_m$.

\subsection{Neighbor Embedding Projection}
\label{subsec:neighbor_embedding_projection}

The Uniform Manifold Approximation and Projection (UMAP)~\citep{mcinnes2018umap}  is a dimensionality reduction technique that aims to preserve the structure of the data. It constructs the k-nearest Neighbors (k-NN) graph in the high-dimensional space and optimizes it to produce a compact, lower-dimensional representation, improving data interpretability while retaining essential characteristics for effective analysis and retrieval.

UMAP's process involves creating a fuzzy simplicial set to model data connectivity in the original space. The method then applies an optimization step that minimizes the cross-entropy divergence between the fuzzy sets in the high- and low-dimensional spaces. This formulation introduces flexibility in how data points are connected, allowing for an adaptive representation of data relationships.

In content-based image retrieval (CBIR), UMAP has proven to be effective in generating more discriminative feature spaces~\citep{leticio2024manifold, kawai2024}. By projecting image features into a lower-dimensional space, the method can improve the distinction between similar images and enhance the overall structure of the retrieval space. Additionally, reducing the dimensionality helps improve computational efficiency by making similarity comparisons faster and more effective.

However, UMAP’s performance is influenced by hyperparameters, such as the number of neighbors (\textit{n\_neighbors}) and the minimum distance (\textit{min\_dist}). Poorly tuned parameters can lead to embeddings that fail to accurately reflect the intrinsic data structure, negatively impacting retrieval quality. Moreover, datasets with varying local densities can suffer from inconsistencies in the projected space, requiring careful parameter selection to balance structure preservation and retrieval effectiveness.

\subsection{Rank-Based Manifold Learning}
\label{subsec:rank_based_manifold}

The application of unsupervised distance learning techniques plays a crucial role in refining similarity relationships for retrieval tasks. In order to simplify access to the implementation of these methods, pyUDLF (a Python framework for unsupervised distance learning methods)~\citep{10.1145/3581783.3613466} provides an interface that facilitates their use and evaluation. In this work, we use pyUDLF to investigate the rank aggregation of UMAP's ranking with three ranking-based methods: LHRR, CPRR, and RFE. Each of these methods refines ranked lists in different ways, allowing a more effective representation of similarity relationships, as described below:

\noindent $\bullet$ \textbf{Log-based Hypergraph of Ranking References (LHRR)}~\citep{pedronette2019multimedia}:
A method that models high-order similarity relationships using hypergraphs. Instead of relying solely on pairwise connections, LHRR represents groups of related elements as hyperedges, capturing more complex structures in the data. A log-based weighting function adjusts the connections, improving retrieval effectiveness.

\noindent $\bullet$ \textbf{Cartesian Product of Ranking References (CPRR)}~\citep{valem2018unsupervised}:
This approach improves similarity estimation by applying Cartesian product operations to neighborhood sets and rank information. By considering pairwise relations weighted according to rank-based information, it generates a refined similarity measure, leading to more effective retrieval results.

\noindent $\bullet$ \textbf{Rank Flow Embedding (RFE)}~\citep{valem2023rank}: 
This method refines similarity relationships by iteratively updating embeddings through a sequence of rank-based operations. It leverages hypergraphs, Cartesian product techniques, and connected components to enhance contextual representation. The resulting embeddings can be used for re-ranking tasks or as improved feature representations.


\subsection{Rank Aggregation}
\label{subsec:rank_agg}

In this work, the Borda Count method is used to aggregate ranked lists. Borda Count~\citep{emerson2013original} is a rank aggregation technique that assigns points to items based on their positions across multiple ranked lists. Specifically, for each ranked list, an item receives a score corresponding to the number of candidates ranked below it. These scores are then summed across all ranked lists, and the final ranking is determined by ordering the items based on their total scores in descending order.

This approach effectively consolidates different ranking perspectives by giving higher weight to consistently high-ranking items while mitigating the impact of outliers. In the context of this work, Borda Count~\citep{emerson2013original} is employed to merge the ranked lists obtained from UMAP-based projections and rank-based manifold learning methods.


\section{Experimental Evaluation}
\label{sec:experimental_eval}

This section presents the experimental evaluation of the proposed approach. Section~\ref{subsec:datasets_and_features} describes the image datasets and feature extraction models used in the experiments. Section~\ref{subsec:experimental_protocol} details the experimental protocol, including parameter settings and implementation aspects. Section~\ref{subsec:results} reports the retrieval results across different datasets and feature representations. Section~\ref{subsec:rank_fusion_comparison} compares the proposed rank aggregation strategy with alternative fusion methods. Section~\ref{subsec:umap_hyperparameter} analyzes the sensitivity of the framework to UMAP hyperparameters, considering both effectiveness and computational cost. Section~\ref{subsec:ablation_study} presents an ablation study to assess the contribution of the aggregation component. Finally, Section~\ref{subsec:statistical_analysis} provides a statistical analysis evaluating the robustness and significance of the obtained results.


\subsection{Datasets and Features}
\label{subsec:datasets_and_features}

 The experimental evaluation was conducted using several well-known public datasets, described below:

\noindent $\bullet$ \textbf{Flowers}~\citep{PaperFlowers}: This dataset contains 1,360 images from 17 distinct flower species, with each species represented by 80 samples.

\noindent $\bullet$ \textbf{Corel5k}~\citep{PaperCorel5k_PR2013}: This dataset features a wide variety of scenes, including images of fireworks, vehicles, microscopic views, tiles, trees, and more. It consists of 50 categories, each containing 100 images.

\noindent $\bullet$ \textbf{Oxford-IIIT Pets}~\citep{parkhi2012cats}: This dataset contains 7,390 images of 37 different pet breeds, including both cats and dogs. Each class contains approximately 200 images, with variations in pose, lighting conditions, and background.

\noindent $\bullet$ \textbf{CUB200-2011}~\citep{wah_cub200_2011}: The Caltech-UCSD Birds (CUB200-2011) dataset contains 11,788 images of 200 bird species. The dataset is widely used for fine-grained visual recognition tasks due to the high visual similarity between classes.

\noindent $\bullet$ \textbf{Dogs}~\citep{KhoslaYaoJayadevaprakashFeiFei_FGVC2011}: Containing a total of 20,580 images, this dataset represents 120 distinct dog breeds from different parts of the world. In contrast to the Flowers and Corel5k datasets, which have a uniform number of images per class, the Dogs dataset presents slight variations in class distribution.

For the experimental evaluation, deep learning features were extracted from three different models: a Convolutional Neural Network (CNN), ResNet152~\citep{paperRESNET}, and two Transformer-based architectures, SwinTf~\citep{paperSWIN-TF} and DinoV2~\citep{oquab2023dinov2}. 
ResNet152 and Swin Transformer were pre-trained on ImageNet~\citep{PaperImageNet}, while DINOv2 was pre-trained on the LVD-142M dataset using self-supervised learning.
Feature representations were extracted from the penultimate pooled layer for ResNet152 and Swin Transformer, and from the final-layer $\texttt{[CLS]}$ token embedding for DINOv2.

More specifically, we employed ResNet152\footnote{\url{https://github.com/Cadene/pretrained-models.pytorch}}, Swin Transformer (Swin-Base, input resolution 224$\times$224, pre-trained on ImageNet-1k)\footnote{\url{https://github.com/microsoft/Swin-Transformer}}, and DINOv2 (ViT-B/14)\footnote{\url{https://github.com/facebookresearch/dinov2}}.

Before feature extraction, all images were converted to RGB, resized to 224$\times$224, and normalized using the standard ImageNet preprocessing scheme. Feature vectors were extracted from the representations prior to the final classification layer, without additional fine-tuning.


\subsection{Experimental Protocol}
\label{subsec:experimental_protocol}

The experimental protocol follows an unsupervised setting, where all images are treated as queries in the Flowers, Corel5k, Pets, CUB200, and Dogs datasets. Euclidean distance was used in all experiments, and the Ball Tree~\citep{PaperExNNAlg_PKDD2007} was employed as the indexing method, considering a ranked list size of 1000 for all datasets.

To evaluate effectiveness, we considered Precision (at dataset-specific depths), and Mean Average Precision (MAP). The retrieval evaluation followed a multi-step process. Initially, retrieval was performed using the original feature representations. Next, UMAP was applied before retrieval to project the embeddings into a lower-dimensional space, leading to improved ranking results. Re-ranking techniques were then applied separately to both the original retrieval results and those obtained with UMAP-reduced features (our initial approach~\citep{kawai2024}). Afterward, a rank aggregation step was performed, combining retrieval results from UMAP-reduced features with the re-ranked lists. Finally, an additional re-rank stage was applied to refine the aggregated lists, employing the same post-processing methods (RFE, CPRR, and LHRR).

Regarding UMAP parameters, we set $n\_components = 2$, $n\_neighbors = 15$, and used Euclidean distance as the metric (default parameters from \textit{umap-learn} library). To ensure reproducibility, we also set the random state of UMAP to $42$. For the RFE, CPRR, and LHRR methods, the default parameters from the pyUDLF~\footnote{\url{https://github.com/UDLF/pyUDLF}} framework were applied, except for the parameter $K$, which was adjusted according to the dataset size. Specifically, we set $K = 80$ for Flowers (1,360 images), $K = 100$ for the intermediate-sized datasets Corel5k (5,000 images) and Pets (7,390 images), and $K = 120$ for the larger datasets CUB200 (11,788 images) and Dogs (20,580 images). Additionally, for all methods, we set the ranked list size to $L = 1000$ and the number of iterations to $T = 2$.

All experiments were conducted on a laptop equipped with an AMD Ryzen 7 5800H CPU, 16GB of RAM, and an NVIDIA GeForce RTX 3050 GPU (4GB VRAM), running Ubuntu 22.04.3 LTS. The implementation was developed in Python 3.12.8, using NumPy 2.1.0, Scikit-learn 1.6.1, umap-learn 0.5.7, PyTorch 2.6.0, and torchvision 0.21.0.

To further support reproducibility, the complete source code for the proposed framework is publicly available at \href{https://github.com/ViniciusAtsushi/Manifold-Rank-Fusion}{https://github.com/ViniciusAtsushi/Manifold-Rank-Fusion}.

\begin{figure*}[ht!]
    \centering
    \includegraphics[width=.95\textwidth]{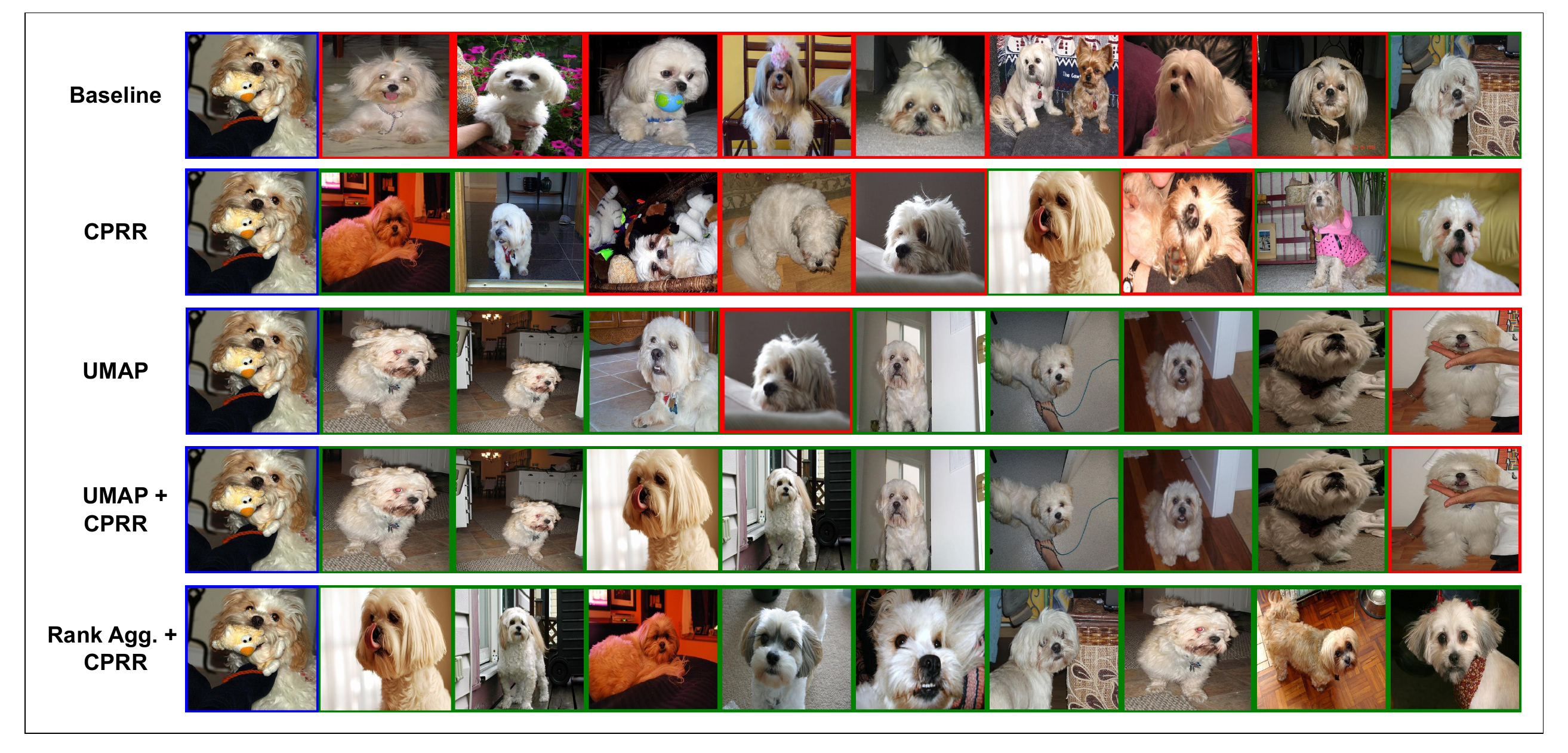}
    \vspace{5mm}
    \caption{Visual Retrieval Results on Dogs dataset for different configurations.}
    \label{fig:dogs_retrieval_visual_rks}
\end{figure*}


\subsection{Retrieval Results}
\label{subsec:results}

This subsection presents the experimental results reporting the effectiveness of the proposed framework across different datasets and feature extraction models. In this sense, Tables~\ref{tab:map_results_flowers} and \ref{tab:p80_results_flowers} present the Mean Average Precision (MAP) and P@80 results for the Flowers dataset, since Flowers contains only 80 images per class. Similarly, the MAP and P@100 results for the Corel5k dataset are detailed in Tables~\ref{tab:map_results_corel5k} and \ref{tab:p100_results_corel5k}. 
The results for the Pets and CUB200 datasets are presented in Tables~\ref{tab:map_results_pets}, \ref{tab:p100_results_pets}, \ref{tab:map_results_cub200}, and \ref{tab:p100_results_cub200}, respectively. Finally, Tables~\ref{tab:map_results_dogs} and \ref{tab:p100_results_dogs} report the MAP and P@100 results for the Dogs dataset.

Additionally, some qualitative results are illustrated through visual retrieval examples presented in Figure~\ref{fig:dogs_retrieval_visual_rks}, which demonstrate the impact of different configurations on retrieval quality.

On the Flowers dataset, the baseline performance for DinoV2 features achieved 98.77\% in MAP indicating a high initial effectiveness due to the relatively simple nature of this dataset with only 80 images per class. In this case, the rank aggregation strategy showed limited gains, as the baseline already approached the upper bound of retrieval accuracy. Notably, 'Re-Rank only' with LHRR reached 100.0\% in MAP. However, for the ResNet152 features, where the baseline MAP was 51.62\%, the proposed approach demonstrated significant improvements.

Analyzing the intermediate configurations, applying 'UMAP only' improved the MAP to 73.32\%, while 'Re-Rank only' with RFE achieved 73.00\%. Combining UMAP + RFE following our initial approach \citep{kawai2024} resulted in a MAP of 75.07\%. These results highlight the progressive enhancements provided by each stage. However, the most substantial gain was achieved using UMAP + Rank Aggregation (using RFE) + Post Re-Rank (using LHRR), where the MAP increased to 78.01\%. This emphasizes the effectiveness of consolidating multiple ranking perspectives through rank aggregation + post re-rank, significantly enhancing retrieval effectiveness by leveraging the complementary information from UMAP's neighbor embedding projections rankings and the contextual adjustments of re-ranking methods.

On the Corel5k dataset, the proposed method showed competitive results. For ResNet152 features, the baseline MAP was 64.50\%, which increased to 88.74\% using UMAP + Rank Aggregation + Post Re-Rank with LHRR and RFE. However, in this scenario, the MAP was slightly lower than the 89.76\% achieved by 'LHRR only'. Despite this, the P@100 metric demonstrated gains (presented in Table~\ref{tab:p100_results_corel5k}), with P@100 improving from 61.29\% in the baseline to 86.08\% using the proposed rank aggregation approach. This suggests that, although our overall MAP was slightly lower than the best configuration, the ranking quality was preserved in the most relevant positions, not only maintaining but reaching the highest retrieval accuracy at the top 100 positions of the ranked list.

Both the Pets and CUB200 datasets present fine-grained retrieval scenarios, where inter-class similarity is high and intra-class variability is significant. In these settings, a similar behavior can be observed for SwinTf features: the highest MAP is achieved by the UMAP-only configuration (84.45\% for Pets and 73.76\% for CUB200), with the aggregation-based strategies achieving close results (84.43\% and 73.47\%, respectively). Nevertheless, the proposed approach remains competitive in terms of top-k precision. For example, on the Pets dataset, the UMAP + Rank Aggregation with CPRR (without post re-ranking) achieves the highest P@100 (87.26\%), while on CUB200 the best P@100 (48.40\%) is obtained using UMAP + Rank Aggregation followed by CPRR post re-ranking. Similar to Corel5k dataset, these results indicate that, even when improvements in global ranking quality (MAP) are limited, the aggregation strategy can still enhance the ordering of the most relevant items.

On the other hand, when considering weaker baseline representations, the benefits of the proposed framework become more evident. For instance, on the CUB200 dataset with ResNet152 features, the baseline MAP is relatively low (22.77\%), likely reflecting the difficulty of fine-grained discrimination in the original feature space. In this case, both UMAP-only (31.29\%) and re-ranking (CPRR: 32.56\%) improve the results, while the best performance is achieved by the full framework (UMAP + CPRR with CPRR post re-ranking), reaching 34.80\% in MAP. This behavior highlights that the effectiveness of the aggregation strategy is more pronounced when the initial representation is less discriminative.

For the Dogs dataset with DinoV2 features, the baseline MAP was 55.18\%, increasing to 66.94\% with UMAP. Combining UMAP with Rank Aggregation (LHRR) further improved the MAP to 68.29\%, demonstrating the benefits of consolidating multiple ranking perspectives. Notably, Rank Aggregation alone achieved the highest MAP, outperforming all configurations, including those with Post Re-Rank. 

It is also important to note that the proposed strategy does not outperform all configurations across every feature extractor. For example, in the Pets dataset with SwinTf features, the highest MAP is achieved using UMAP alone. This behavior suggests that, depending on the feature representation and dataset characteristics, the complementary information between projection-based and rank-based strategies may not always be fully exploited. Additionally, in some cases the aggregation step alone already provides most of the achievable improvement, while the inclusion of an additional post re-ranking stage may produce diminishing returns, similarly to the effect observed when increasing the number of refinement iterations in rank-based re-ranking methods. Nevertheless, in several scenarios the combination of these strategies leads to improved retrieval effectiveness, particularly when the baseline performance is relatively low, where the benefits of aggregation become more pronounced.

Overall, the results indicate that the proposed rank aggregation strategy can improve retrieval effectiveness in several scenarios. By combining rankings obtained from projection-based and rank-based manifold learning methods, the approach is able to exploit complementary information between these strategies. While improvements are not observed uniformly across all datasets and feature extractors, the results demonstrate that the proposed aggregation scheme can provide competitive performance and, in many cases, lead to gains in retrieval effectiveness.


\subsection{Comparison with Alternative Rank Aggregation Methods}
\label{subsec:rank_fusion_comparison}

In addition to the Borda Count strategy adopted in the proposed framework, we also evaluated alternative rank fusion methods commonly used in information retrieval. In particular, we considered Reciprocal Rank Fusion (RRF)~\citep{cormack2009reciprocal} and CombSUM~\citep{lee1995combining}. For a fair comparison, these methods were applied to the same ranked lists used in the Borda-based aggregation, combining the rankings obtained from the UMAP projection and from the re-ranking stage.

The RRF method was implemented following its standard formulation, where each item receives a contribution of $1/(k+r)$ from each ranked list, with $r$ denoting the rank position and $k=60$. In contrast, CombSUM typically operates on similarity scores rather than rank positions. However, the ranked lists generated by the re-ranking methods (LHRR, CPRR, and RFE) are produced through manifold-based analysis of neighborhood relationships and do not directly provide geometric distance values that could be normalized and aggregated with the distances obtained from the UMAP projection. For this reason, we adopted a positional score transformation for CombSUM, assigning each item a score of $1/(r+1)$ according to its rank position $r$ in each list. These scores were then summed across rankings to produce the final aggregated order.

The comparison was conducted on the Dogs dataset using the same three feature extractors employed throughout the experimental evaluation: DINOv2, ResNet152, and Swin Transformer. This setup allows a consistent comparison between aggregation strategies while maintaining the same retrieval pipeline and ranked lists used in the proposed framework.

Tables \ref{tab:map_dogs_fusion_comparison} and \ref{tab:p100_dogs_fusion_comparison} present the MAP and P@100 results, respectively, obtained when comparing the three aggregation methods. The results show that the Borda Count adopted in the proposed framework provides competitive performance when compared with the alternative aggregation methods. In most configurations, Borda Count achieves equal or higher retrieval effectiveness than both RRF and CombSUM. This behavior suggests that the positional voting scheme employed by Borda Count is well suited for combining rankings generated from heterogeneous stages of the pipeline, such as projection-based rankings and manifold-based re-ranking methods.

Additionally, the results indicate that the performance differences between aggregation strategies may vary depending on the feature representation. Nevertheless, across the evaluated configurations, the Borda Count aggregation remains competitive and frequently provides the best results among the evaluated fusion strategies.


\subsection{UMAP: Hyperparameter Evaluation}
\label{subsec:umap_hyperparameter}

Since dimensionality reduction plays a central role in the proposed framework, we analyze how retrieval effectiveness behaves under different UMAP configurations. In particular, the quality of the low-dimensional embedding may be influenced by hyperparameters controlling the projection structure, such as the number of projection dimensions, the neighborhood size used to approximate the manifold, and the minimum distance between embedded points. To evaluate the sensitivity of the framework to these parameters, we conducted experiments varying the main UMAP hyperparameters while keeping the remaining components of the pipeline unchanged. The evaluation was performed on the Dogs dataset, the largest dataset considered in this work, using the same three feature extractors evaluated throughout the paper (DINOv2, ResNet152, and Swin Transformer). In addition to retrieval effectiveness measured by Mean Average Precision (MAP), we also analyzed the computational cost of the main stages of the framework.

In this evaluation, the rankings obtained from the UMAP projection were aggregated with the rankings produced by the CPRR re-ranking method using the Borda Count strategy. The post re-ranking stage was not applied in order to allow a clearer assessment of the impact of the UMAP projection parameters on the aggregated ranking, since an additional refinement stage could partially mask the effect of the projection variations.

The sensitivity analysis was performed by varying one parameter at a time while keeping the remaining parameters fixed at the default values of the \textit{umap-learn} library. In this manner, the evaluated configurations were defined as follows:
\begin{itemize}
\item $n\_components \in \{2, 8, 16, 32\}$
\item $n\_neighbors \in \{5, 15, 30, 50\}$
\item $min\_dist \in \{0.0, 0.1, 0.3, 0.5\}$
\end{itemize}

\begin{figure*}[t]
\centering

\begin{subfigure}[t]{0.32\textwidth}
\centering
\includegraphics[width=\linewidth]{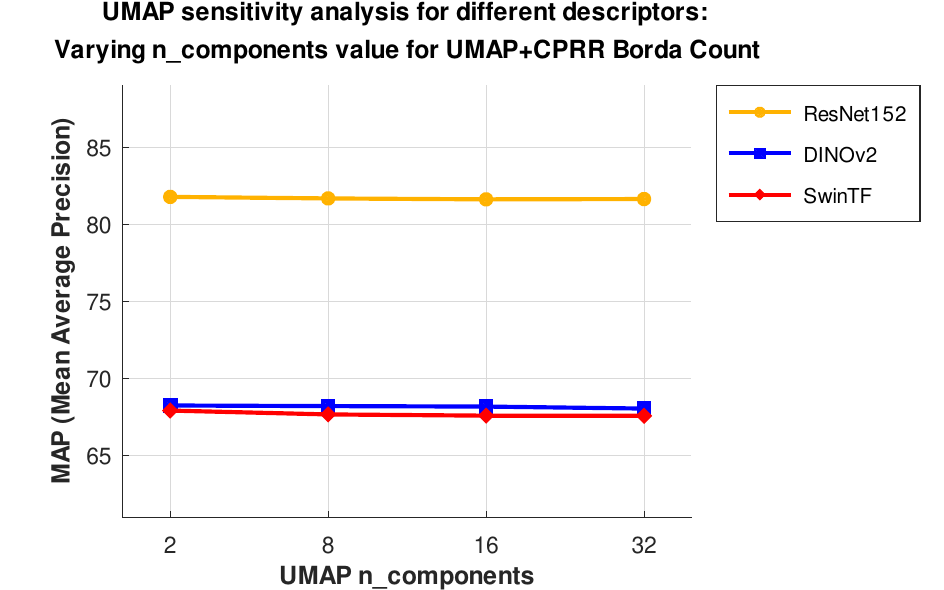}
\caption{Variation of $n\_components$.}
\label{fig:umap_n_components}
\end{subfigure}
\hfill
\begin{subfigure}[t]{0.32\textwidth}
\centering
\includegraphics[width=\linewidth]{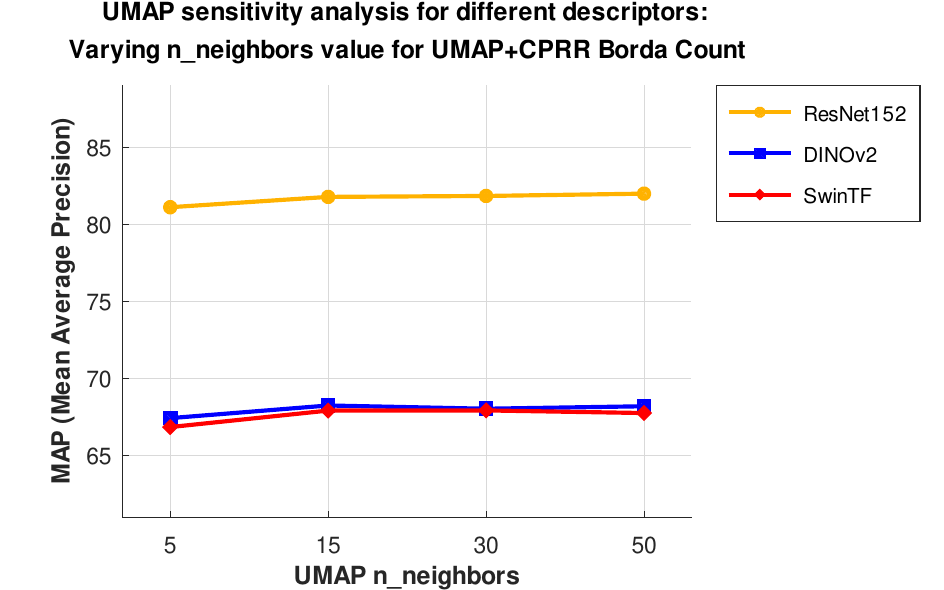}
\caption{Variation of $n\_neighbors$.}
\label{fig:umap_n_neighbors}
\end{subfigure}
\hfill
\begin{subfigure}[t]{0.32\textwidth}
\centering
\includegraphics[width=\linewidth]{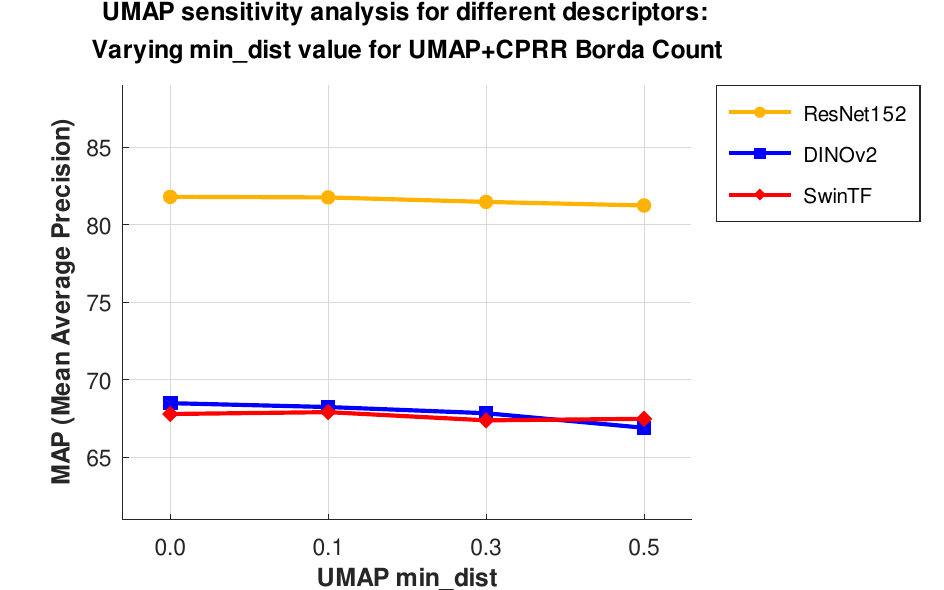}
\caption{Variation of $min\_dist$.}
\label{fig:umap_min_dist}
\end{subfigure}
\vspace{3mm}
\caption{Sensitivity analysis of the main UMAP hyperparameters on the Dogs dataset. Each plot reports the MAP obtained when varying a single parameter while keeping the others fixed at their default values. Results are shown for the three feature extractors evaluated in this work.}
\label{fig:umap_sensitivity}
\end{figure*}

Figures \ref{fig:umap_n_components}, \ref{fig:umap_n_neighbors}, and \ref{fig:umap_min_dist} present the MAP values obtained when varying each parameter for the three feature extractors. Each plot reports the retrieval performance after aggregating the rankings from the UMAP projection and the CPRR re-ranking method using Borda Count. Overall, the results show that retrieval performance remains stable across the evaluated configurations. Although small variations can be observed depending on the descriptor and the selected parameter value, the MAP differences remain limited across the tested settings. In the context of an unsupervised retrieval scenario, this robustness is desirable since parameter selection cannot rely on labeled validation data.

Additionally, Table \ref{tab:table_umap_runtimes} reports the average runtime for the main stages of the framework, including the UMAP projection, the ranking computation in the projected space, and the rank aggregation step. For each UMAP parameter configuration, the projection, ranking and aggregation stages were executed three times per descriptor, and the reported values correspond to the average runtime obtained per descriptor followed by the average across the three evaluated descriptors. Although CPRR is deterministic and a single execution would be sufficient, it was also executed three times per descriptor to report a comparable average runtime. This analysis allows evaluating the trade-off between retrieval effectiveness and computational cost under different UMAP parameter settings.

\begin{table}[h]
\centering
\caption{Average execution time (seconds) for the main stages of the framework across different UMAP parameter configurations (Dogs dataset).}
\resizebox{0.49\textwidth}{!}{
\begin{tabular}{lccc}
\hline
Configuration & UMAP Projection & Ranking (RK) & Borda Aggregation \\
\hline
$n\_components=2$  & 10.41 $\pm$ 0.80 s & 3.18 $\pm$ 0.24 s & 10.84 $\pm$ 0.25 s \\
$n\_components=32$ & 14.71 $\pm$ 0.67 s & 7.29 $\pm$ 0.53 s & 10.61 $\pm$ 0.09 s \\
$n\_neighbors=5$   & 6.51 $\pm$ 0.26 s & 3.26 $\pm$ 0.01 s & 10.70 $\pm$ 0.25 s \\
$n\_neighbors=50$  & 22.63 $\pm$ 3.45 s & 3.33 $\pm$ 0.35 s & 10.77 $\pm$ 0.11 s \\
$min\_dist=0.0$   & 10.02 $\pm$ 0.93 s & 3.19 $\pm$ 0.21 s & 10.56 $\pm$ 0.14 s \\
$min\_dist=0.5$  & 10.04 $\pm$ 0.97 s &  3.43 $\pm$ 0.06 s & 10.73 $\pm$ 0.23 s \\
\hline
CPRR (re-rank only) & \multicolumn{3}{c}{5.92 $\pm$  0.32 s} \\
\hline
\end{tabular}
}
\label{tab:table_umap_runtimes}
\end{table}

In this manner, increasing the number of projection dimensions ($n\_components$) leads to higher runtime during the rank computation stage, since the Euclidean distance calculations are performed in a higher-dimensional space. In contrast, variations in the neighborhood parameter ($n\_neighbors$) mainly affect the runtime of the UMAP projection step, as larger neighborhoods require additional computations during the construction of the manifold approximation. On the other hand, varying the $min\_dist$ parameter does not produce significant changes in the runtime of the evaluated stages. Finally, the runtime of the Borda Count aggregation remains approximately constant across all configurations, since this step depends only on the aggregation of ranked lists and is independent of the dimensionality of the embedding.


\subsection{Ablation Study}
\label{subsec:ablation_study}

In order to better understand the contribution of each component of the proposed framework, we conducted an ablation study evaluating a simplified configuration of the pipeline. In this experiment, the ranked lists obtained from the original feature space and from the UMAP projection were directly combined using the Borda Count aggregation strategy, without applying any re-ranking method.

This configuration, referred to as \textit{Borda Only (Original + UMAP)}, allows isolating the impact of the rank aggregation step from the additional refinement provided by the manifold-based re-ranking methods. In this way, the experiment evaluates whether the aggregation of projection-based and original rankings alone is sufficient to improve retrieval effectiveness.

The evaluation was conducted on the Dogs dataset using the same three feature extractors considered throughout the experimental evaluation (DINOv2, ResNet152, and Swin Transformer). Table~\ref{tab:ablation_borda_only} presents the MAP and P@100 results.

The results indicate that combining the rankings from the original feature space and the UMAP projection without the re-ranking stage does not consistently improve retrieval performance. For multiple features, the simplified configuration produces lower effectiveness compared to the full framework. These findings suggest that the improvements observed in the proposed approach are not solely due to the aggregation of rankings from different representations, but rather from the complementary interaction between projection-based representations and manifold-based re-ranking refinement.

\begin{table}[t]
\centering
\caption{Ablation study on the Dogs dataset. Comparison between baseline configurations, the simplified \textit{Borda Only (Original + UMAP)} aggregation, and the best configuration of the proposed framework. The best results are shown in \textbf{bold}, while the second-best results are \underline{underlined}.}
\resizebox{0.49\textwidth}{!}{
\begin{tabular}{lcccccc}
\hline
\multirow{2}{*}{Method} & \multicolumn{2}{c}{DinoV2} & \multicolumn{2}{c}{ResNet152} & \multicolumn{2}{c}{SwinTf} \\
 & MAP & P@100 & MAP & P@100 & MAP & P@100 \\
\hline
Baseline            & 55.18 &  63.57 & 63.73 & 72.74 & 45.53 & 59.83 \\
UMAP Only           & \underline{66.94} & \underline{71.85} & \underline{80.77} & 83.40 & \underline{69.65} & 74.08 \\
Re-Rank Only (best) & 65.41 & 71.17 & 79.97 & \textbf{85.61} & 61.69 & \underline{75.60} \\
Borda Only (Original + UMAP) & 65.86 & 71.73 & 77.80 & 82.78 & 63.65 & 71.83 \\
\hline
Proposed Framework (Best Config.) & \textbf{68.29} & \textbf{73.03} & \textbf{82.60}
& \underline{85.41} & \textbf{71.74} & \textbf{75.68} \\
\hline
\end{tabular}
}
\label{tab:ablation_borda_only}
\end{table}


\subsection{Statistical Analysis}
\label{subsec:statistical_analysis}

The experimental protocol described in Section \ref{subsec:experimental_protocol} employs fixed UMAP parameters and a predefined random state to ensure reproducibility of the reported results. However, since UMAP relies on stochastic optimization during the embedding process, different random seeds may lead to slightly different projections.

To assess the stability of the proposed framework under different executions of the projection step, we conducted an additional robustness analysis using multiple runs of UMAP. Specifically, the projection was executed five times using different random seeds. For each run, retrieval results were computed following the same experimental protocol described in Section \ref{subsec:experimental_protocol}. It is important to note that UMAP is the only stochastic component of the proposed pipeline. The subsequent steps, including rank aggregation and the re-ranking methods, are fully deterministic and therefore produce identical results given the same input ranked lists.

This analysis was conducted on the Dogs dataset, considering the three feature extractors evaluated in this work (ResNet152, SwinTf, and DINOv2). Tables \ref{tab:map_results_dogs_est_map} and \ref{tab:map_results_dogs_est_p100} report the Mean Average Precision (MAP) and Precision@100 results, respectively, presented as mean $\pm$ standard deviation across five independent runs of UMAP with different random seeds.

In addition to the variance analysis, we conducted statistical significance tests to evaluate whether the observed differences between methods are consistent across queries. As a reference baseline, we considered the strongest configuration that does not rely on the proposed aggregation strategy. Based on the MAP results reported in Table \ref{tab:map_results_dogs_est_map}, this baseline corresponds to the \textit{UMAP Only} setting. For each feature extractor, the best result obtained with the proposed approach (results in \textbf{bold}) was compared against this baseline.

The statistical comparison was performed using the paired Wilcoxon signed-rank test over the Average Precision (AP) values computed for each query. Due to the stochastic nature of UMAP, the AP values were first averaged across multiple runs with different random seeds, and the tests were conducted using these per-query averaged values. Furthermore, to quantify the magnitude of the observed differences, we report the matched-pairs Rank-Biserial Correlation (RBC), the effect size measure associated with the Wilcoxon test.

The statistical comparison yielded the following results:

\begin{itemize}
    \item \textbf{ResNet152}: The proposed configuration (82.39\% MAP) achieved a higher score than the baseline (80.59\% MAP), with statistical significance ($p < 0.001$) and an effect size of $\mathrm{RBC}=0.6107$.
    
    \item \textbf{SwinTf}: The proposed configuration (71.77\% MAP) achieved a higher score than the baseline (69.70\% MAP), with statistical significance ($p < 0.001$) and an effect size of $\mathrm{RBC}=0.5719$.
    
    \item \textbf{DINOv2}: The proposed configuration (68.22\% MAP) achieved a higher score than the baseline (66.87\% MAP), with statistical significance ($p < 0.001$) and an effect size of $\mathrm{RBC}=0.4921$.
\end{itemize}

The Wilcoxon test ($p < 0.001$) indicates that the observed differences between the compared configurations are statistically significant. The positive RBC values further indicate that the proposed configuration tends to produce higher Average Precision values across queries compared to the baseline.


\section{Conclusion}
\label{sec:conclusion}

This paper presented a rank aggregation framework that integrates neighbor embedding projections with rank-based manifold learning to enhance content-based image retrieval. The proposed approach combines UMAP-based dimensionality reduction with multiple re-ranking methods, aggregating the resulting ranked lists using the Borda Count strategy followed by an optional post re-ranking stage.

The experimental evaluation across multiple datasets and feature extraction models demonstrated that the proposed aggregation strategy can improve retrieval effectiveness in several scenarios. In particular, noticeable gains were observed when the baseline feature representation did not already achieve very high retrieval performance, where the complementary information provided by projection-based and rank-based manifold learning methods becomes more relevant. Additionally, the proposed approach often improves the quality of the top-ranked positions, which is particularly important for practical retrieval applications.

At the same time, the results also show that the benefits of the proposed pipeline may vary depending on the dataset and feature representation, especially in cases where the baseline already achieves very high performance. These observations highlight the importance of combining multiple retrieval perspectives while carefully considering the characteristics of the underlying feature space.

In this manner, future research directions may include exploring alternative dimensionality reduction techniques, feature representations, and rank aggregation strategies to further optimize retrieval performance. Additionally, investigating the influence of other hyperparameter settings, such as different distance metrics in UMAP and correlation metrics in re-ranking methods, could lead to more adaptive and effective retrieval outcomes.


\section*{Declarations}

\begin{contributions}
\textbf{Vinicius Atsushi} contributed to the Conceptualization, Data curation, Software, Methodology, Resources, Validation, Visualization, Writing – original draft, and Writing – review \& editing. 
\textbf{Gustavo Leticio} contributed to the Data curation, Software, Resources, Visualization, and Writing – review \& editing. 
\textbf{Lucas Valem} contributed to the Data curation, Software, Supervision, and Writing – review \& editing. 
\textbf{Daniel Pedronette} contributed to the Supervision, Conceptualization, Funding acquisition, Project administration, and Writing – review \& editing.
All authors read and approved the final manuscript.
\end{contributions}

\begin{interests}
The authors declare that they have no competing interests.
\end{interests}

\begin{acknowledgements}
The authors gratefully acknowledge the Department of Statistics, Applied Mathematics and Computing (DEMAC) at São Paulo State University (UNESP) for providing the hardware resources required to conduct the experiments reported in this work.
Additionally, this work was supported by grants from FAPESP, CNPq, Petrobras, CAPES, and the University of São Paulo (USP), whose financial support is gratefully acknowledged.
\end{acknowledgements}

\begin{funding}
The authors are grateful to São Paulo Research Foundation - FAPESP (grants \#2024/04890-5, \#2025/07171-2, and \#2025/10602-5), Brazilian National Council for Scientific and Technological Development - CNPq (grants \#313193/2023-1 and \#422667/2021-8),  Petrobras (grant \#2023/00095-3), and the University of São Paulo (PRPI Ordinance No. 1032, New Faculty Support Program) for financial support. This study was financed in part by the Coordenação de Aperfeiçoamento de Pessoal de Nível Superior - Brasil (CAPES).
\end{funding}

\begin{materials}
The datasets used in this work are publicly available at the following sources: Flowers (\href{https://www.robots.ox.ac.uk/~vgg/data/flowers/}{oxford17flowers}), Corel5k (\href{https://www.kaggle.com/datasets/parhamsalar/corel5k}{corel5k}), Pets (\href{https://www.robots.ox.ac.uk/~vgg/data/pets/}{pets}), CUB200 (\href{https://www.vision.caltech.edu/datasets/cub_200_2011/}{cub200}), and Dogs (\href{http://vision.stanford.edu/aditya86/ImageNetDogs/}{StanfordDogs}). The re-ranking methods were executed using a publicly accessible framework (\href{https://github.com/UDLF/pyUDLF}{pyUDLF}), and UMAP implementation used umap-learn library (\href{https://umap-learn.readthedocs.io/en/latest/}{umap}) ensuring reproducibility of the experimental results. The complete source code for the proposed framework is available at \href{https://github.com/ViniciusAtsushi/Manifold-Rank-Fusion}{https://github.com/ViniciusAtsushi/Manifold-Rank-Fusion}.
\end{materials}

\bibliographystyle{apalike-sol}
\bibliography{refs}

\newpage
\clearpage

\begin{table*}[ht!]
    \centering
    \caption{\textbf{Mean Average Precision (MAP)} Results for Different Configurations on Flowers Dataset.}
    \begin{tabular}{|c|lll|ccc|}
        \hline
        \multirow{2}{*}{\centering \textbf{Method}} & \multicolumn{3}{c}{\textbf{Method Specification}}   & \multicolumn{3}{|c|}{\textbf{Feature}} \\ \cline{2-5} \cline{6-7}
        & \textbf{Projection} & \textbf{Re-Rank} & \textbf{Post Re-Rank}   & \textbf{DinoV2} & \textbf{ResNet152} & \textbf{SwinTf} \\
        \hline
        \multirow{1}{*}{None - Baseline} 
        & -     & -     & -        & 98.77\%  & 51.62\%  & 92.91\%  \\ 
        \hline
        \multirow{1}{*}{UMAP Only} 
        & UMAP  & -     & -       & 98.64 \%  & 73.32 \% & 99.24 \% \\ 
        \hline
        \multirow{3}{*}{Re-Rank Only} 
            & -     & CPRR  & -       & 99.85 \%  & 73.10 \%  & 98.55 \%  \\
            & -     & RFE   & -       & 99.95 \%  & 73.00 \%  & 99.15 \%  \\
            & -     & LHRR  & -       & \textbf{100.0 \%}  & 75.34 \%  & 99.66 \%  \\
        \hline
        \multirow{3}{*}{\shortstack{UMAP + Re-Rank \\~\cite{kawai2024}}}
            & UMAP  & CPRR  & -        & 98.61 \%  & 73.99 \% & 99.37 \% \\
            & UMAP  & RFE   & -        & 98.64 \%  & 75.07 \% & 99.31 \% \\
            & UMAP  & LHRR  & -        & 98.59 \%  & 74.60 \% & 99.36 \% \\
        \hline
        \multirow{3}{*}{\shortstack{UMAP Rank Aggregation\\ (UMAP + Re-Rank)}}
            & UMAP  & CPRR  & -     & 98.82 \% & 74.84 \% & 98.62 \% \\
            & UMAP  & RFE   & -     & 98.82 \% & 74.97 \% & 99.29 \% \\
            & UMAP  & LHRR  & -     & 98.81 \% & 76.08 \% & 99.52 \% \\
        \hline
        \multirow{9}{*}{\shortstack{UMAP Rank Aggregation\\ (UMAP + Re-Rank) \\ + Post Re-Rank}}
            & UMAP  & CPRR  & CPRR     & 98.77 \% & 76.32 \% & 99.66 \% \\
            & UMAP  & CPRR  & RFE      & 98.82 \% & 77.61 \% & 99.66 \% \\
            & UMAP  & CPRR  & LHRR     & 98.77 \% & 76.87 \% & 99.69 \% \\
            & UMAP  & RFE   & CPRR     & 98.79 \% & 77.63 \% & 99.55 \% \\
            & UMAP  & RFE   & RFE      & 98.92 \% & 76.36 \% & 99.35 \% \\
            & UMAP  & RFE   & LHRR     & 98.74 \% & \textbf{78.01 \%} & 99.58 \% \\
            & UMAP  & LHRR  & CPRR     & 98.78 \% & 76.97 \% & 99.70 \% \\
            & UMAP  & LHRR  & RFE      & 98.81 \% & 77.56 \% & 99.64 \% \\
            & UMAP  & LHRR  & LHRR     & 98.73 \% & 77.40 \% & \textbf{99.73 \%} \\
        \hline
    \end{tabular}
    \label{tab:map_results_flowers}
\end{table*}

\begin{table*}[ht!]
    \centering
    \caption{\textbf{Precision at 80 (P@80)} Results for Different Configurations on Flowers Dataset.}
    \begin{tabular}{|c|lll|ccc|}
        \hline
        \multirow{2}{*}{\centering \textbf{Method}} & \multicolumn{3}{c}{\textbf{Method Specification}}   & \multicolumn{3}{|c|}{\textbf{Feature}} \\ \cline{2-5} \cline{6-7}
        & \textbf{Projection} & \textbf{Re-Rank} & \textbf{Post Re-Rank}   & \textbf{DinoV2} & \textbf{ResNet152} & \textbf{SwinTf} \\
        \hline
        \multirow{1}{*}{None - Baseline} 
        & -     & -     & -        & 97.27 \%  & 49.19 \%  & 90.59 \%  \\ 
        \hline
        \multirow{1}{*}{UMAP Only} 
        & UMAP  & -     & -       & 98.60 \%  & 69.27 \% & 98.88 \% \\ 
        \hline
        \multirow{3}{*}{Re-Rank Only} 
            & -     & CPRR  & -       & 99.73 \%  & 69.26 \% & 98.48 \% \\
            & -     & RFE   & -       & 99.84 \%  & 69.20 \% & 98.80 \% \\
            & -     & LHRR  & -       & \textbf{100.0} \%  & 70.53 \% & 99.51 \% \\
        \hline
        \multirow{3}{*}{\shortstack{UMAP + Re-Rank \\~\cite{kawai2024}}}
            & UMAP  & CPRR  & -     & 98.44 \% & 69.87 \% & 98.95 \% \\
            & UMAP  & RFE   & -     & 98.60 \% & 70.23 \% & 98.92 \% \\
            & UMAP  & LHRR  & -     & 98.40 \% & 69.84 \% & 98.96 \% \\
        \hline
        \multirow{3}{*}{\shortstack{UMAP Rank Aggregation\\ (UMAP + Re-Rank)}}
            & UMAP  & CPRR  & -     & 98.60 \% & 70.15 \% & 98.23 \% \\
            & UMAP  & RFE   & -     & 98.60 \% & 70.57 \% & 98.92 \% \\
            & UMAP  & LHRR  & -     & 98.60 \% & 71.13 \% & 99.19 \% \\
        \hline
        \multirow{9}{*}{\shortstack{UMAP Rank Aggregation\\ (UMAP + Re-Rank) \\ + Post Re-Rank}}
            & UMAP  & CPRR  & CPRR     & 98.52 \% & 70.92 \% & 99.41 \% \\
            & UMAP  & CPRR  & RFE      & 98.60 \% & 72.62 \% & 99.50 \% \\
            & UMAP  & CPRR  & LHRR     & 98.47 \% & 71.53 \% & \textbf{99.55} \% \\
            & UMAP  & RFE   & CPRR     & 98.53 \% & 72.48 \% & 99.04 \% \\
            & UMAP  & RFE   & RFE      & 98.60 \% & 72.62 \% & 98.91 \% \\
            & UMAP  & RFE   & LHRR     & 98.41 \% & \textbf{73.02} \% & 99.07 \% \\
            & UMAP  & LHRR  & CPRR     & 98.47 \% & 71.65 \% & 99.34 \% \\
            & UMAP  & LHRR  & RFE      & 98.60 \% & 72.76 \% & 99.25 \% \\
            & UMAP  & LHRR  & LHRR     & 98.40 \% & 71.95 \% & 99.40 \% \\
        \hline
    \end{tabular}
    \label{tab:p80_results_flowers}
\end{table*}


\begin{table*}[ht!]
    \centering
    \caption{\textbf{Mean Average Precision (MAP)} Results for Different Configurations on Corel5k Dataset.}
    \begin{tabular}{|c|lll|ccc|}
        \hline
        \multirow{2}{*}{\centering \textbf{Method}} & \multicolumn{3}{c}{\textbf{Method Specification}}   & \multicolumn{3}{|c|}{\textbf{Feature}} \\ \cline{2-5} \cline{6-7}
        & \textbf{Projection} & \textbf{Re-Rank} & \textbf{Post Re-Rank}   & \textbf{DinoV2} & \textbf{ResNet152} & \textbf{SwinTf} \\
        \hline
        \multirow{1}{*}{None - Baseline} 
        & -     & -     & -        & 81.28 \% & 64.50 \% & 73.92 \% \\ 
        \hline
        \multirow{1}{*}{UMAP Only} 
        & UMAP  & -     & -       & 89.52 \% & 85.13 \% & 93.92 \% \\ 
        \hline
        \multirow{3}{*}{Re-Rank Only} 
            & -     & CPRR  & -       & 90.86 \% & 83.50 \% & 87.46 \% \\
            & -     & RFE   & -       & 94.51 \% & 87.83 \% & 95.91 \% \\
            & -     & LHRR  & -       & \textbf{95.08 \%} & \textbf{89.76 \%} & 96.93 \% \\
        \hline
        \multirow{3}{*}{\shortstack{UMAP + Re-Rank \\~\cite{kawai2024}}}
            & UMAP  & CPRR  & -        & 90.14 \% & 85.64 \% & 94.12 \% \\
            & UMAP  & RFE   & -        & 90.26 \% & 86.53 \% & 94.33 \% \\
            & UMAP  & LHRR  & -        & 89.52 \% & 86.07 \% & 94.22 \% \\
        \hline
        \multirow{3}{*}{\shortstack{UMAP Rank Aggregation\\ (UMAP + Re-Rank)}}
            & UMAP  & CPRR  & -     & 90.05 \% & 84.45 \% & 90.20 \% \\
            & UMAP  & RFE   & -     & 92.31 \% & 87.21 \% & 95.32 \% \\
            & UMAP  & LHRR  & -     & 92.63 \% & 88.04 \% & 95.69 \% \\
        \hline
        \multirow{9}{*}{\shortstack{UMAP Rank Aggregation\\ (UMAP + Re-Rank) \\ + Post Re-Rank}}
            & UMAP  & CPRR  & CPRR     & 93.43 \% & 87.95 \% & 95.39 \% \\
            & UMAP  & CPRR  & RFE      & 93.86 \% & 88.63 \% & 95.55 \% \\
            & UMAP  & CPRR  & LHRR     & 94.10 \% & 88.15 \% & 95.41 \% \\
            & UMAP  & RFE   & CPRR     & 94.37 \% & 88.62 \% & 96.13 \% \\
            & UMAP  & RFE   & RFE      & 93.76 \% & 87.90 \% & 95.82 \% \\
            & UMAP  & RFE   & LHRR     & 94.60 \% & 87.98 \% & 95.99 \% \\
            & UMAP  & LHRR  & CPRR     & 94.20 \% & 88.62 \% & 96.63 \% \\
            & UMAP  & LHRR  & RFE      & 94.66 \% & 88.74 \% & \textbf{97.72 \%} \\
            & UMAP  & LHRR  & LHRR     & 94.38 \% & 88.51 \% & 97.09 \% \\
        \hline
    \end{tabular}
    \label{tab:map_results_corel5k}
\end{table*}

\begin{table*}[ht!]
    \centering
    \caption{\textbf{Precision at 100 (P@100)} Results for Different Configurations on Corel5k Dataset.}
    \begin{tabular}{|c|lll|ccc|}
        \hline
        \multirow{2}{*}{\centering \textbf{Method}} & \multicolumn{3}{c}{\textbf{Method Specification}}   & \multicolumn{3}{|c|}{\textbf{Feature}} \\ \cline{2-5} \cline{6-7}
        & \textbf{Projection} & \textbf{Re-Rank} & \textbf{Post Re-Rank}   & \textbf{DinoV2} & \textbf{ResNet152} & \textbf{SwinTf} \\
        \hline
        \multirow{1}{*}{None - Baseline} 
        & -     & -     & -        & 77.81 \%  & 61.29 \%  & 71.27 \%  \\ 
        \hline
        \multirow{1}{*}{UMAP Only} 
        & UMAP  & -     & -       & 87.72 \%  & 83.15 \% & 93.15 \% \\ 
        \hline
        \multirow{3}{*}{Re-Rank Only} 
            & -     & CPRR  & -       & 89.44 \%  & 81.08 \%  & 87.42 \%  \\
            & -     & RFE   & -       & 92.35 \%  & 83.85 \%  & 94.33 \% \\
            & -     & LHRR  & -       & \textbf{92.84} \%  & 85.90 \%  & 95.28 \%  \\
        \hline
        \multirow{3}{*}{\shortstack{UMAP + Re-Rank \\~\cite{kawai2024}}}
            & UMAP  & CPRR  & -        & 88.40 \%  & 83.73 \% & 93.46 \% \\
            & UMAP  & RFE   & -        & 88.73 \%  & 84.64 \% & 93.75 \% \\
            & UMAP  & LHRR  & -        & 88.75 \%  & 83.93 \% & 93.53 \% \\
        \hline
        \multirow{3}{*}{\shortstack{UMAP Rank Aggregation\\ (UMAP + Re-Rank)}}
            & UMAP  & CPRR  & -     & 86.30 \% & 81.07 \% & 85.61 \% \\
            & UMAP  & RFE   & -     & 89.27 \% & 83.93 \% & 93.49 \% \\
            & UMAP  & LHRR  & -     & 89.65 \% & 84.68 \% & 93.62 \% \\
        \hline
        \multirow{9}{*}{\shortstack{UMAP Rank Aggregation\\ (UMAP + Re-Rank) \\ + Post Re-Rank}}
            & UMAP  & CPRR  & CPRR     & 91.00 \% & 84.91 \% & 93.75 \% \\
            & UMAP  & CPRR  & RFE      & 91.57 \% & 85.77 \% & 94.33 \% \\
            & UMAP  & CPRR  & LHRR     & 92.00 \% & 85.52 \% & 94.22 \% \\
            & UMAP  & RFE   & CPRR     & 92.12 \% & 85.56 \% & 94.92 \% \\
            & UMAP  & RFE   & RFE      & 91.52 \% & 85.47 \% & 94.52 \% \\
            & UMAP  & RFE   & LHRR     & \textbf{92.84} \% & 85.38 \% & 94.99 \% \\
            & UMAP  & LHRR  & CPRR     & 91.95 \% & 85.47 \% & 94.87 \% \\
            & UMAP  & LHRR  & RFE      & 92.40 \% & \textbf{86.08} \% & \textbf{96.71} \% \\
            & UMAP  & LHRR  & LHRR     & 92.54 \% & 85.80 \% & 95.73 \% \\
        \hline
    \end{tabular}
    \label{tab:p100_results_corel5k}
\end{table*}


\begin{table*}[ht!]
    \centering
    \caption{\textbf{Mean Average Precision (MAP)} Results for Different Configurations on Pets Dataset.}
    \resizebox{0.99\textwidth}{!}{
    \begin{tabular}{|c|lll|ccc|}
        \hline
        \multirow{2}{*}{\centering \textbf{Method}} & \multicolumn{3}{c}{\textbf{Method Specification}}   & \multicolumn{3}{|c|}{\textbf{Feature}} \\ \cline{2-5} \cline{6-7}
        & \textbf{Projection} & \textbf{Re-Rank} & \textbf{Post Re-Rank}   & \textbf{DinoV2} & \textbf{ResNet152} & \textbf{SwinTf} \\
        \hline
        \multirow{1}{*}{None - Baseline} 
        & -     & -     & -        & 79.65 \% & 68.28 \% & 59.06 \% \\ 
        \hline
        \multirow{1}{*}{UMAP Only} 
        & UMAP  & -     & -       & 87.25 \% & 84.84 \% & \textbf{84.45} \% \\ 
        \hline
        \multirow{3}{*}{Re-Rank Only} 
            & -     & CPRR  & -       & 86.08 \% & 81.86 \% & 75.78 \% \\
            & -     & RFE   & -       & 80.89 \% & 76.65 \% & 65.82 \% \\
            & -     & LHRR  & -       & 85.15 \% & 82.34 \% & 74.85 \%  \\
        \hline
        \multirow{3}{*}{\shortstack{UMAP + Re-Rank \\\textit{Kawai et al.} [2024]}}
            & UMAP  & CPRR  & -        & 86.33 \% & 83.47 \% & 83.33 \% \\
            & UMAP  & RFE   & -        & 87.22 \% & 83.93 \% & 83.46 \% \\
            & UMAP  & LHRR  & -        & 86.59 \% & 83.49 \% & 83.42 \% \\
        \hline
        \multirow{3}{*}{\shortstack{UMAP Rank Aggregation\\ (UMAP + Re-Rank)}}
            & UMAP  & CPRR  & -     & \textbf{88.34} \% & 84.80 \% & 81.78 \% \\
            & UMAP  & RFE   & -     & 86.81 \% & 83.45 \% & 78.84 \% \\
            & UMAP  & LHRR  & -     & 88.14 \% & \textbf{85.27} \% & 82.20 \% \\
        \hline
        \multirow{9}{*}{\shortstack{UMAP Rank Aggregation\\ (UMAP + Re-Rank) \\ + Post Re-Rank}}
            & UMAP  & CPRR  & CPRR     & 87.39 \% & 84.72 \% & 84.43 \% \\
            & UMAP  & CPRR  & RFE      & 87.65 \% & 85.18 \% & 83.95 \% \\
            & UMAP  & CPRR  & LHRR     & 87.51 \% & 84.39 \% & 84.40 \% \\
            & UMAP  & RFE   & CPRR     & 86.03 \% & 83.75 \% & 83.19 \% \\
            & UMAP  & RFE   & RFE      & 85.07 \% & 79.41 \% & 76.19 \% \\
            & UMAP  & RFE   & LHRR     & 86.27 \% & 81.43 \% & 81.02 \%  \\
            & UMAP  & LHRR  & CPRR     & 86.90 \% & 84.25 \% & 84.17 \% \\
            & UMAP  & LHRR  & RFE      & 86.84 \% & 83.25 \% & 82.34 \% \\
            & UMAP  & LHRR  & LHRR     & 87.05 \% & 82.98 \% & 82.64 \% \\
        \hline
    \end{tabular}
    }
    \label{tab:map_results_pets}
\end{table*}

\begin{table*}[ht!]
    \centering
    \caption{\textbf{Precision at 100 (P@100)} Results for Different Configurations on Pets Dataset.}
    \resizebox{0.99\textwidth}{!}{
    \begin{tabular}{|c|lll|ccc|}
        \hline
        \multirow{2}{*}{\centering \textbf{Method}} & \multicolumn{3}{c}{\textbf{Method Specification}}   & \multicolumn{3}{|c|}{\textbf{Feature}} \\ \cline{2-5} \cline{6-7}
        & \textbf{Projection} & \textbf{Re-Rank} & \textbf{Post Re-Rank}   & \textbf{DinoV2} & \textbf{ResNet152} & \textbf{SwinTf} \\
        \hline
        \multirow{1}{*}{None - Baseline} 
        & -     & -     & -        & 84.81 \%  & 78.58 \%  & 74.20 \%  \\ 
        \hline
        \multirow{1}{*}{UMAP Only} 
        & UMAP  & -     & -       & 89.13 \%  & 87.18 \% & 86.01 \% \\ 
        \hline
        \multirow{3}{*}{Re-Rank Only} 
            & -     & CPRR  & -       & 88.34 \%  & 87.26 \%  & 86.58 \%  \\
            & -     & RFE   & -       & 84.32 \%  & 83.30 \%  & 77.34 \%  \\
            & -     & LHRR  & -       & 87.27 \%  & 86.07 \%  & 80.47 \%  \\
        \hline
        \multirow{3}{*}{\shortstack{UMAP + Re-Rank \\\textit{Kawai et al.} [2024]}}
            & UMAP  & CPRR  & -        & 88.48 \%  & 86.16 \% & 85.14 \% \\
            & UMAP  & RFE   & -        & 88.88 \%  & 86.23 \% & 85.01 \% \\
            & UMAP  & LHRR  & -        & 88.51 \%  & 85.92 \% & 85.08 \% \\
        \hline
        \multirow{3}{*}{\shortstack{UMAP Rank Aggregation\\ (UMAP + Re-Rank)}}
            & UMAP  & CPRR  & -     & \textbf{89.85} \% & \textbf{87.99} \% & \textbf{87.26} \% \\
            & UMAP  & RFE   & -     & 88.91 \% & 87.35 \% & 84.77 \% \\
            & UMAP  & LHRR  & -     & 89.69 \% & 87.49 \% & 84.53 \% \\
        \hline
        \multirow{9}{*}{\shortstack{UMAP Rank Aggregation\\ (UMAP + Re-Rank) \\ + Post Re-Rank}}
            & UMAP  & CPRR  & CPRR     & 89.08 \% & 86.75 \% & 86.15 \% \\
            & UMAP  & CPRR  & RFE      & 89.20 \% & 87.33 \% & 86.10 \% \\
            & UMAP  & CPRR  & LHRR     & 89.08 \% & 86.85 \% & 86.26 \% \\
            & UMAP  & RFE   & CPRR     & 88.20 \% & 85.95 \% & 84.69 \% \\
            & UMAP  & RFE   & RFE      & 87.66 \% & 83.41 \% & 81.34 \% \\
            & UMAP  & RFE   & LHRR     & 88.08 \% & 85.05 \% & 83.20 \% \\
            & UMAP  & LHRR  & CPRR     & 89.05 \% & 86.68 \% & 85.78 \% \\
            & UMAP  & LHRR  & RFE      & 88.87 \% & 86.57 \% & 84.91 \% \\
            & UMAP  & LHRR  & LHRR     & 88.88 \% & 86.35 \% & 85.34 \% \\
        \hline
    \end{tabular}
    }
    \label{tab:p100_results_pets}
\end{table*}


\begin{table*}[ht!]
    \centering
    \caption{\textbf{Mean Average Precision (MAP)} Results for Different Configurations on CUB200 Dataset.}
    \resizebox{0.99\textwidth}{!}{
    \begin{tabular}{|c|lll|ccc|}
        \hline
        \multirow{2}{*}{\centering \textbf{Method}} & \multicolumn{3}{c}{\textbf{Method Specification}}   & \multicolumn{3}{|c|}{\textbf{Feature}} \\ \cline{2-5} \cline{6-7}
        & \textbf{Projection} & \textbf{Re-Rank} & \textbf{Post Re-Rank}   & \textbf{DinoV2} & \textbf{ResNet152} & \textbf{SwinTf} \\
        \hline
        \multirow{1}{*}{None - Baseline} 
        & -     & -     & -        & 65.54 \% & 22.77 \% & 58.27 \% \\ 
        \hline
        \multirow{1}{*}{UMAP Only} 
        & UMAP  & -     & -       & 73.97 \% & 31.29 \% & \textbf{73.76} \% \\ 
        \hline
        \multirow{3}{*}{Re-Rank Only} 
            & -     & CPRR  & -       & 71.49 \% & 32.56 \% & 67.82 \% \\
            & -     & RFE   & -       & 66.74 \% & 30.86 \% & 60.66 \% \\
            & -     & LHRR  & -       & 68.14 \% & 31.39 \% & 64.24 \%  \\
        \hline
        \multirow{3}{*}{\shortstack{UMAP + Re-Rank \\\textit{Kawai et al.} [2024]}}
            & UMAP  & CPRR  & -        & 72.36 \% & 30.77 \% & 72.97 \% \\
            & UMAP  & RFE   & -        & 73.21 
\% & 31.12 \% & 73.51 \% \\
            & UMAP  & LHRR  & -        & 71.29 \% & 30.48 \% & 72.20 \% \\
        \hline
        \multirow{3}{*}{\shortstack{UMAP Rank Aggregation\\ (UMAP + Re-Rank)}}
            & UMAP  & CPRR  & -     & \textbf{74.87} \% & 34.48 \% & 73.02 \% \\
            & UMAP  & RFE   & -     & 72.85 \% & 33.88 \% & 69.88 \% \\
            & UMAP  & LHRR  & -     & 73.61 \% & 34.00 \% & 72.21 \% \\
        \hline
        \multirow{9}{*}{\shortstack{UMAP Rank Aggregation\\ (UMAP + Re-Rank) \\ + Post Re-Rank}}
            & UMAP  & CPRR  & CPRR     & 73.56 \% & \textbf{34.80} \% & 73.47 \% \\
            & UMAP  & CPRR  & RFE      & 72.35 \% & 33.07 \% & 71.36 \% \\
            & UMAP  & CPRR  & LHRR     & 71.43 \% & 32.95 \% & 71.34 \% \\
            & UMAP  & RFE   & CPRR     & 71.92 \% & 34.36 \% & 71.57 \% \\
            & UMAP  & RFE   & RFE      & 68.77 \% & 30.57 \% & 65.47 \% \\
            & UMAP  & RFE   & LHRR     & 68.79 \% & 31.98 \% & 67.91 \% \\
            & UMAP  & LHRR  & CPRR     & 72.56 \% & 34.33 \% & 72.86 \% \\
            & UMAP  & LHRR  & RFE      & 69.93 \% & 31.54 \% & 69.44 \% \\
            & UMAP  & LHRR  & LHRR     & 70.01 \% & 32.25 \% & 70.22 \% \\
        \hline
    \end{tabular}
    }
    \label{tab:map_results_cub200}
\end{table*}

\begin{table*}[ht!]
    \centering
    \caption{\textbf{Precision at 100 (P@100)} Results for Different Configurations on CUB200 Dataset.}
    \resizebox{0.99\textwidth}{!}{
    \begin{tabular}{|c|lll|ccc|}
        \hline
        \multirow{2}{*}{\centering \textbf{Method}} & \multicolumn{3}{c}{\textbf{Method Specification}}   & \multicolumn{3}{|c|}{\textbf{Feature}} \\ \cline{2-5} \cline{6-7}
        & \textbf{Projection} & \textbf{Re-Rank} & \textbf{Post Re-Rank}   & \textbf{DinoV2} & \textbf{ResNet152} & \textbf{SwinTf} \\
        \hline
        \multirow{1}{*}{None - Baseline} 
        & -     & -     & -        & 43.21 \%  & 19.67 \%  & 39.90 \%  \\ 
        \hline
        \multirow{1}{*}{UMAP Only} 
        & UMAP  & -     & -       & 47.56 \%  & 24.20 \% & 48.14 \% \\ 
        \hline
        \multirow{3}{*}{Re-Rank Only} 
            & -     & CPRR  & -       & 47.06 \%  & 26.33 \%  & 46.30 \%  \\
            & -     & RFE   & -       & 44.23 \%  & 24.19 \%  & 42.09 \%  \\
            & -     & LHRR  & -       & 45.56 \%  & 24.96 \%  & 44.01 \%  \\
        \hline
        \multirow{3}{*}{\shortstack{UMAP + Re-Rank \\\textit{Kawai et al.} [2024]}}
            & UMAP  & CPRR  & -        & 47.39 \%  & 24.17 \% & 48.03 \% \\
            & UMAP  & RFE   & -        & 47.35 \%  & 24.25 \% & 48.05 \% \\
            & UMAP  & LHRR  & -        & 47.11 \%  & 24.09 \% & 47.80 \% \\
        \hline
        \multirow{3}{*}{\shortstack{UMAP Rank Aggregation\\ (UMAP + Re-Rank)}}
            & UMAP  & CPRR  & -     & \textbf{47.78} \% & 25.80 \% & 47.43 \% \\
            & UMAP  & RFE   & -     & 46.46 \% & 25.49 \% & 45.70 \% \\
            & UMAP  & LHRR  & -     & 47.15 \% & 25.62 \% & 47.16 \% \\
        \hline
        \multirow{9}{*}{\shortstack{UMAP Rank Aggregation\\ (UMAP + Re-Rank) \\ + Post Re-Rank}}
            & UMAP  & CPRR  & CPRR     & 47.66 \% & \textbf{26.55} \% & \textbf{48.40} \% \\
            & UMAP  & CPRR  & RFE      & 47.15 \% & 25.51 \% & 47.54 \% \\
            & UMAP  & CPRR  & LHRR     & 47.30 \% & 25.69 \% & 47.91 \% \\
            & UMAP  & RFE   & CPRR     & 46.94 \% & 26.20 \% & 47.63 \% \\
            & UMAP  & RFE   & RFE      & 45.63 \% & 24.07 \% & 44.35 \% \\
            & UMAP  & RFE   & LHRR     & 46.23 \% & 25.02 \% & 46.33 \% \\
            & UMAP  & LHRR  & CPRR     & 47.39 \% & 26.26 \% & 48.21 \% \\
            & UMAP  & LHRR  & RFE      & 46.45 \% & 24.69 \% & 46.88 \% \\
            & UMAP  & LHRR  & LHRR     & 46.90 \% & 25.33 \% & 47.63 \% \\
        \hline
    \end{tabular}
    }
    \label{tab:p100_results_cub200}
\end{table*}


\begin{table*}[ht!]
    \centering
    \caption{\textbf{Mean Average Precision (MAP)} Results for Different Configurations on Dogs Dataset.}
    \begin{tabular}{|c|lll|ccc|}
        \hline
        \multirow{2}{*}{\centering \textbf{Method}} & \multicolumn{3}{c}{\textbf{Method Specification}}   & \multicolumn{3}{|c|}{\textbf{Feature}} \\ \cline{2-5} \cline{6-7}
        & \textbf{Projection} & \textbf{Re-Rank} & \textbf{Post Re-Rank}   & \textbf{DinoV2} & \textbf{ResNet152} & \textbf{SwinTf} \\
        \hline
        \multirow{1}{*}{None - Baseline} 
        & -     & -     & -        & 55.18 \% & 63.73 \% & 45.53 \% \\ 
        \hline
        \multirow{1}{*}{UMAP Only} 
        & UMAP  & -     & -       & 66.94 \% & 80.77 \% & 69.65 \% \\ 
        \hline
        \multirow{3}{*}{Re-Rank Only} 
            & -     & CPRR  & -       & 64.92 \% & 79.80 \% & 61.69 \% \\
            & -     & RFE   & -       & 62.06 \% & 72.99 \% & 50.19 \% \\
            & -     & LHRR  & -       & 65.41 \% & 79.97 \% & 61.39 \%  \\
        \hline
        \multirow{3}{*}{\shortstack{UMAP + Re-Rank \\~\cite{kawai2024}}}
            & UMAP  & CPRR  & -        & 66.21 \% & 80.20 \% & 68.89 \% \\
            & UMAP  & RFE   & -        & 66.14 \% & 80.43 \% & 69.46 \% \\
            & UMAP  & LHRR  & -        & 66.15 \% & 80.39 \% & 69.09 \% \\
        \hline
        \multirow{3}{*}{\shortstack{UMAP Rank Aggregation\\ (UMAP + Re-Rank)}}
            & UMAP  & CPRR  & -     & 68.24 \% & 81.77 \% & 67.91 \% \\
            & UMAP  & RFE   & -     & 66.97 \% & 79.19 \% & 63.35 \% \\
            & UMAP  & LHRR  & -     & \textbf{68.29 \%} & 82.00 \% & 68.17 \% \\
        \hline
        \multirow{9}{*}{\shortstack{UMAP Rank Aggregation\\ (UMAP + Re-Rank) \\ + Post Re-Rank}}
            & UMAP  & CPRR  & CPRR     & 67.75 \% & \textbf{82.60 \%} & \textbf{71.74 \%} \\
            & UMAP  & CPRR  & RFE      & 67.30 \% & 82.28 \% & 70.34 \% \\
            & UMAP  & CPRR  & LHRR     & 66.46 \% & 82.06 \% & 70.88 \% \\
            & UMAP  & RFE   & CPRR     & 67.08 \% & 81.47 \% & 68.36 \% \\
            & UMAP  & RFE   & RFE      & 62.81 \% & 74.13 \% & 57.04 \% \\
            & UMAP  & RFE   & LHRR     & 64.39 \% & 78.87 \% & 64.03 \%  \\
            & UMAP  & LHRR  & CPRR     & 67.62 \% & 82.47 \% & 71.10 \% \\
            & UMAP  & LHRR  & RFE      & 66.09 \% & 80.99 \% & 67.30 \% \\
            & UMAP  & LHRR  & LHRR     & 65.66 \% & 81.87 \% & 69.16 \% \\
        \hline
    \end{tabular}
    \label{tab:map_results_dogs}
\end{table*}

\begin{table*}[ht!]
    \centering
    \caption{\textbf{Precision at 100 (P@100)} Results for Different Configurations on Dogs Dataset.}
    \begin{tabular}{|c|lll|ccc|}
        \hline
        \multirow{2}{*}{\centering \textbf{Method}} & \multicolumn{3}{c}{\textbf{Method Specification}}   & \multicolumn{3}{|c|}{\textbf{Feature}} \\ \cline{2-5} \cline{6-7}
        & \textbf{Projection} & \textbf{Re-Rank} & \textbf{Post Re-Rank}   & \textbf{DinoV2} & \textbf{ResNet152} & \textbf{SwinTf} \\
        \hline
        \multirow{1}{*}{None - Baseline} 
        & -     & -     & -        & 63.57 \%  & 72.74 \%  & 59.83 \%  \\ 
        \hline
        \multirow{1}{*}{UMAP Only} 
        & UMAP  & -     & -       & 71.85 \%  & 83.84 \% & 74.08 \% \\ 
        \hline
        \multirow{3}{*}{Re-Rank Only} 
            & -     & CPRR  & -       & 71.17 \%  & \textbf{85.61 \%}  & 75.60 \%  \\
            & -     & RFE   & -       & 68.52 \%  & 81.24 \%  & 63.46 \%  \\
            & -     & LHRR  & -       & 70.53 \%  & 83.83 \%  & 69.40 \%  \\
        \hline
        \multirow{3}{*}{\shortstack{UMAP + Re-Rank \\~\cite{kawai2024}}}
            & UMAP  & CPRR  & -        & 70.74 \%  & 82.91 \% & 72.99 \% \\
            & UMAP  & RFE   & -        & 70.71 \%  & 83.16 \% & 73.65 \% \\
            & UMAP  & LHRR  & -        & 70.58 \%  & 82.88 \% & 73.08 \% \\
        \hline
        \multirow{3}{*}{\shortstack{UMAP Rank Aggregation\\ (UMAP + Re-Rank)}}
            & UMAP  & CPRR  & -     & \textbf{73.03 \%} & 85.41 \% & 75.62 \% \\
            & UMAP  & RFE   & -     & 72.16 \% & 84.54 \% & 71.83 \% \\
            & UMAP  & LHRR  & -     & 72.84 \% & 84.76 \% & 73.29 \% \\
        \hline
        \multirow{9}{*}{\shortstack{UMAP Rank Aggregation\\ (UMAP + Re-Rank) \\ + Post Re-Rank}}
            & UMAP  & CPRR  & CPRR     & 71.52 \% & 84.43 \% & 75.31 \% \\
            & UMAP  & CPRR  & RFE      & 71.40 \% & 84.77 \% & 75.66 \% \\
            & UMAP  & CPRR  & LHRR     & 71.34 \% & 84.64 \% & \textbf{75.68 \%} \\
            & UMAP  & RFE   & CPRR     & 71.25 \% & 83.58 \% & 72.39 \% \\
            & UMAP  & RFE   & RFE      & 69.34 \% & 79.66 \% & 65.95 \% \\
            & UMAP  & RFE   & LHRR     & 70.55 \% & 82.36 \% & 69.54 \% \\
            & UMAP  & LHRR  & CPRR     & 71.68 \% & 84.60 \% & 75.03 \% \\
            & UMAP  & LHRR  & RFE      & 71.20 \% & 84.22 \% & 73.70 \% \\
            & UMAP  & LHRR  & LHRR     & 71.54 \% & 84.83 \% & 75.14 \% \\
        \hline
    \end{tabular}
    \label{tab:p100_results_dogs}
\end{table*}


\begin{table*}[ht!]
    \centering
    \caption{\textbf{Mean Average Precision (MAP)} Results for Different Configurations on Dogs Dataset: \textbf{Comparing BordaCount, CombSUM, and RRF}.}
    \resizebox{0.99\textwidth}{!}{
    \begin{tabular}{|c|c|c|l|l|ccc|}
        \hline
        \multirow{2}{*}{\centering \textbf{Method}} & \multicolumn{4}{c}{\textbf{Method Specification}}   & \multicolumn{3}{|c|}{\textbf{Feature}} \\ \cline{2-5} \cline{5-8}
        & \textbf{Rank Aggregation} & \textbf{Projection} & \textbf{Re-Rank} & \textbf{Post Re-Rank}   & \textbf{DinoV2} & \textbf{ResNet152} & \textbf{SwinTf} \\
        \hline

        \multirow{9}{*}{\shortstack{UMAP Rank Aggregation\\ (UMAP + Re-Rank)}}
          & \multirow{3}{*}{\shortstack{BordaCount}} & \multirow{3}{*}{\shortstack{UMAP}}  
                 & CPRR  & -     & 68.24 \% & 81.77 \% & 67.91 \% \\
          &  &   & RFE   & -     & 66.97 \% & 79.19 \% & 63.35 \% \\
          &  &   & LHRR  & -     & \textbf{68.29 \%} & 82.00 \% & 68.17 \% \\
          \cline{2-8} 
          & \multirow{3}{*}{\shortstack{CombSUM}} & \multirow{3}{*}{\shortstack{UMAP}}  
                 & CPRR  & -     & 67.67 \% & 82.33 \% & 70.17 \% \\
          &  &   & RFE   & -     & 66.70 \% & 80.07 \% & 65.30 \% \\
          &  &   & LHRR  & -     & 67.58 \% & 81.57 \% & 68.28 \% \\
          \cline{2-8} 
          & \multirow{3}{*}{\shortstack{RRF}} & \multirow{3}{*}{\shortstack{UMAP}}  
                 & CPRR  & -     & 68.07 \% & 82.37 \% & 69.99 \% \\
          &  &   & RFE   & -     & 67.19 \% & 80.36 \% & 65.70 \% \\
          &  &   & LHRR  & -     & 68.06 \% & 81.93 \% & 68.78 \% \\
        \hline
        \hline
        \multirow{27}{*}{\shortstack{UMAP Rank Aggregation\\ (UMAP + Re-Rank) \\ + Post Re-Rank}}
          & \multirow{9}{*}{\shortstack{BordaCount}} & \multirow{9}{*}{\shortstack{UMAP}}  
                 & CPRR  & CPRR     & 67.75 \% & \textbf{82.60 \%} & \textbf{71.74 \%} \\
          &  &   & CPRR  & RFE      & 67.30 \% & 82.28 \% & 70.34 \% \\
          &  &   & CPRR  & LHRR     & 66.46 \% & 82.06 \% & 70.88 \% \\
          &  &   & RFE   & CPRR     & 67.08 \% & 81.47 \% & 68.36 \% \\
          &  &   & RFE   & RFE      & 62.81 \% & 74.13 \% & 57.04 \% \\
          &  &   & RFE   & LHRR     & 64.39 \% & 78.87 \% & 64.03 \% \\
          &  &   & LHRR  & CPRR     & 67.62 \% & 82.47 \% & 71.10 \% \\
          &  &   & LHRR  & RFE      & 66.09 \% & 80.99 \% & 67.30 \% \\
          &  &   & LHRR  & LHRR     & 65.66 \% & 81.87 \% & 69.16 \% \\
          \cline{2-8} 
          & \multirow{9}{*}{\shortstack{CombSUM}} & \multirow{9}{*}{\shortstack{UMAP}} 
                 & CPRR  & CPRR     & 67.09 \% & 82.37 \% & 71.35 \% \\
          &  &   & CPRR  & RFE      & 65.54 \% & 81.90 \% & 71.10 \% \\
          &  &   & CPRR  & LHRR     & 65.39 \% & 81.64 \% & 70.35 \% \\
          &  &   & RFE   & CPRR     & 66.82 \% & 81.88 \% & 70.29 \% \\
          &  &   & RFE   & RFE      & 64.07 \% & 80.35 \% & 68.97 \% \\
          &  &   & RFE   & LHRR     & 64.42 \% & 80.40 \% & 69.08 \% \\
          &  &   & LHRR  & CPRR     & 66.94 \% & 82.25 \% & 71.14 \% \\
          &  &   & LHRR  & RFE      & 64.80 \% & 81.12 \% & 69.78 \% \\
          &  &   & LHRR  & LHRR     & 64.72 \% & 81.18 \% & 69.60 \% \\
          \cline{2-8} 
          & \multirow{9}{*}{\shortstack{RRF}} & \multirow{9}{*}{\shortstack{UMAP}} 
                 & CPRR  & CPRR     & 67.35 \% & 82.45 \% & 71.48 \% \\
          &  &   & CPRR  & RFE      & 66.28 \% & 82.26 \% & 70.89 \% \\
          &  &   & CPRR  & LHRR     & 65.76 \% & 81.89 \% & 70.68 \% \\
          &  &   & RFE   & CPRR     & 67.06 \% & 81.91 \% & 70.22 \% \\
          &  &   & RFE   & RFE      & 64.41 \% & 79.27 \% & 66.31 \% \\
          &  &   & RFE   & LHRR     & 64.81 \% & 80.57 \% & 69.01 \% \\
          &  &   & LHRR  & CPRR     & 67.19 \% & 82.32 \% & 71.25 \% \\
          &  &   & LHRR  & RFE      & 65.48 \% & 81.48 \% & 69.81 \% \\
          &  &   & LHRR  & LHRR     & 65.28 \% & 81.55 \% & 69.81 \% \\
        \hline
    \end{tabular}
    }
    \label{tab:map_dogs_fusion_comparison}
\end{table*}

\begin{table*}[ht!]
    \centering
    \caption{\textbf{Precision at 100 (P@100)} Results for Different Configurations on Dogs Dataset: \textbf{Comparing BordaCount, CombSUM, and RRF}.}
    \resizebox{0.99\textwidth}{!}{
    \begin{tabular}{|c|c|c|l|l|ccc|}
        \hline
        \multirow{2}{*}{\centering \textbf{Method}} & \multicolumn{4}{c}{\textbf{Method Specification}}   & \multicolumn{3}{|c|}{\textbf{Feature}} \\ \cline{2-5} \cline{5-8}
        & \textbf{Rank Aggregation} & \textbf{Projection} & \textbf{Re-Rank} & \textbf{Post Re-Rank}   & \textbf{DinoV2} & \textbf{ResNet152} & \textbf{SwinTf} \\
        \hline

        \multirow{9}{*}{\shortstack{UMAP Rank Aggregation\\ (UMAP + Re-Rank)}}
          & \multirow{3}{*}{\shortstack{BordaCount}} & \multirow{3}{*}{\shortstack{UMAP}}  
                 & CPRR  & -     & \textbf{73.03 \%} & \textbf{85.41 \%} & 75.62 \% \\
          &  &   & RFE   & -     & 72.16 \% & 84.54 \% & 71.83 \% \\
          &  &   & LHRR  & -     & 72.84 \% & 84.76 \% & 73.29 \% \\
          \cline{2-8} 
          & \multirow{3}{*}{\shortstack{CombSUM}} & \multirow{3}{*}{\shortstack{UMAP}}  
                 & CPRR  & -     & 72.17 \% & 85.10 \% & 75.62 \% \\\
          &  &   & RFE   & -     & 71.21 \% & 83.44 \% & 70.70 \% \\
          &  &   & LHRR  & -     & 71.92 \% & 84.28 \% & 72.88 \% \\
          \cline{2-8} 
          & \multirow{3}{*}{\shortstack{RRF}} & \multirow{3}{*}{\shortstack{UMAP}}  
                 & CPRR  & -     & 72.54 \% & 85.24 \% & \textbf{75.72 \%} \\
          &  &   & RFE   & -     & 71.72 \% & 83.96 \% & 71.40 \% \\\
          &  &   & LHRR  & -     & 72.34 \% & 84.56 \% & 73.35 \% \\
        \hline
        \hline
        \multirow{27}{*}{\shortstack{UMAP Rank Aggregation\\ (UMAP + Re-Rank) \\ + Post Re-Rank}}
          & \multirow{9}{*}{\shortstack{BordaCount}} & \multirow{9}{*}{\shortstack{UMAP}}  
                 & CPRR  & CPRR     & 71.52 \% & 84.43 \% & 75.31 \% \\
          &  &   & CPRR  & RFE      & 71.40 \% & 84.77 \% & 75.66 \% \\
          &  &   & CPRR  & LHRR     & 71.34 \% & 84.64 \% & 75.68 \% \\
          &  &   & RFE   & CPRR     & 71.25 \% & 83.58 \% & 72.39 \% \\
          &  &   & RFE   & RFE      & 69.34 \% & 79.66 \% & 65.95 \% \\
          &  &   & RFE   & LHRR     & 70.55 \% & 82.36 \% & 69.54 \% \\
          &  &   & LHRR  & CPRR     & 71.68 \% & 84.60 \% & 75.03 \% \\
          &  &   & LHRR  & RFE      & 71.20 \% & 84.22 \% & 73.70 \% \\
          &  &   & LHRR  & LHRR     & 71.54 \% & 84.83 \% & 75.14 \% \\
          \cline{2-8} 
          & \multirow{9}{*}{\shortstack{CombSUM}} & \multirow{9}{*}{\shortstack{UMAP}} 
                 & CPRR  & CPRR     & 71.05 \% & 84.25 \% & 74.87 \% \\
          &  &   & CPRR  & RFE      & 70.53 \% & 84.04 \% & 75.04 \% \\
          &  &   & CPRR  & LHRR     & 70.84 \% & 84.27 \% & 75.12 \% \\
          &  &   & RFE   & CPRR     & 71.11 \% & 84.13 \% & 74.31 \% \\
          &  &   & RFE   & RFE      & 70.28 \% & 83.44 \% & 74.27 \% \\
          &  &   & RFE   & LHRR     & 70.69 \% & 83.91 \% & 74.61 \% \\
          &  &   & LHRR  & CPRR     & 71.18 \% & 84.48 \% & 75.00 \% \\
          &  &   & LHRR  & RFE      & 70.37 \% & 83.69 \% & 74.76 \% \\
          &  &   & LHRR  & LHRR     & 70.70 \% & 84.12 \% & 75.12 \% \\
          \cline{2-8} 
          & \multirow{9}{*}{\shortstack{RRF}} & \multirow{9}{*}{\shortstack{UMAP}} 
                 & CPRR  & CPRR     & 71.21 \% & 84.24 \% & 74.97 \% \\
          &  &   & CPRR  & RFE      & 71.04 \% & 84.43 \% & 75.48 \% \\
          &  &   & CPRR  & LHRR     & 71.11 \% & 84.40 \% & 75.42 \% \\
          &  &   & RFE   & CPRR     & 71.35 \% & 84.10 \% & 74.20 \% \\
          &  &   & RFE   & RFE      & 70.43 \% & 83.02 \% & 72.60 \% \\
          &  &   & RFE   & LHRR     & 70.93 \% & 84.03 \% & 74.43 \% \\
          &  &   & LHRR  & CPRR     & 71.39 \% & 84.52 \% & 75.10 \% \\
          &  &   & LHRR  & RFE      & 70.78 \% & 84.12 \% & 75.02 \% \\
          &  &   & LHRR  & LHRR     & 71.11 \% & 84.39 \% & 75.42 \% \\
        \hline
    \end{tabular}
    }
    \label{tab:p100_dogs_fusion_comparison}
\end{table*}


\begin{table*}[ht!]
    \centering
    \caption{\textbf{Mean Average Precision (MAP)} results on the Dogs dataset across five independent executions varying the stochastic component of the framework (UMAP projection). The table reports the mean and standard deviation obtained for each configuration.}
    \resizebox{0.99\textwidth}{!}{
    \begin{tabular}{|c|lll|ccc|}
        \hline
        \multirow{2}{*}{\centering \textbf{Method}} & \multicolumn{3}{c}{\textbf{Method Specification}}   & \multicolumn{3}{|c|}{\textbf{Feature}} \\ \cline{2-5} \cline{6-7}
        & \textbf{Projection} & \textbf{Re-Rank} & \textbf{Post Re-Rank}   & \textbf{DinoV2} & \textbf{ResNet152} & \textbf{SwinTf} \\
        \hline
        \multirow{1}{*}{None - Baseline} 
        & -     & -     & -        &  55.18 \% & 63.73 \% & 45.53  \% \\ 
        \hline
        \multirow{1}{*}{UMAP Only} 
        & UMAP  & -     & -       &66.87 ± 0.17  \% &80.59 ± 0.20  \% &69.70 ± 0.16  \% \\ 
        \hline
        \multirow{3}{*}{Re-Rank Only} 
            & -     & CPRR  & -       & 64.92 \% & 79.80 \% & 61.69 \% \\
            & -     & RFE   & -       & 62.06 \% & 72.99 \% & 50.19 \% \\
            & -     & LHRR  & -       & 65.41 \% & 79.97 \% & 61.39 \%  \\
        \hline
        \multirow{3}{*}{\shortstack{UMAP + Re-Rank \\\cite{kawai2024}}}
            & UMAP  & CPRR  & -        & 66.06 ± 0.23 \% & 79.88 ± 0.20 \% & 68.89 ± 0.21 \% \\
            & UMAP  & RFE   & -        & 66.17 ± 0.22 \% & 80.13 ± 0.24 \% & 69.51 ± 0.24 \% \\
            & UMAP  & LHRR  & -        & 66.06 ± 0.21 \% & 79.99 ± 0.19 \% & 69.06 ± 0.24 \% \\
        \hline
        \multirow{3}{*}{\shortstack{UMAP Rank Aggregation\\ (UMAP + Re-Rank)}}
            & UMAP  & CPRR  & -     & 68.10 $\pm$ 0.09   \% & 81.64 $\pm$ 0.10  \% & 67.74 $\pm$ 0.10  \% \\
            & UMAP  & RFE   & -     & 66.89 $\pm$ 0.09   \% & 78.92 $\pm$ 0.07  \% & 63.21 $\pm$ 0.13  \% \\
            & UMAP  & LHRR  & -     & \textbf{68.22 $\pm$ 0.11   \%} & 81.89 $\pm$ 0.11  \% & 68.14 $\pm$ 0.12  \% \\
        \hline
        \multirow{9}{*}{\shortstack{UMAP Rank Aggregation\\ (UMAP + Re-Rank) \\ + Post Re-Rank}}
            & UMAP  & CPRR  & CPRR     & 67.51 $\pm$ 0.18  \% & \textbf{82.39 $\pm$ 0.13 \%} & \textbf{71.77 $\pm$ 0.17 \%} \\
            & UMAP  & CPRR  & RFE      & 66.96 $\pm$ 0.10  \% & 81.20 $\pm$ 0.15 \% & 68.28 $\pm$ 0.15 \% \\
            & UMAP  & CPRR  & LHRR     & 67.45 $\pm$ 0.12  \% & 82.26 $\pm$ 0.14 \% & 71.17 $\pm$ 0.17 \% \\
            & UMAP  & RFE   & CPRR     & 67.06 $\pm$ 0.18  \% & 82.05 $\pm$ 0.18 \% & 70.29 $\pm$ 0.22 \% \\
            & UMAP  & RFE   & RFE      & 62.61 $\pm$ 0.16  \% & 73.82 $\pm$ 0.15 \% & 56.71 $\pm$ 0.32 \% \\
            & UMAP  & RFE   & LHRR     & 65.65 $\pm$ 0.12  \% & 80.73 $\pm$ 0.15 \% & 67.43 $\pm$ 0.22 \%  \\
            & UMAP  & LHRR  & CPRR     & 66.13 $\pm$ 0.20  \% & 81.70 $\pm$ 0.19 \% & 71.23 $\pm$ 0.16 \% \\
            & UMAP  & LHRR  & RFE      & 64.18 $\pm$ 0.16  \% & 78.57 $\pm$ 0.15 \% & 63.82 $\pm$ 0.30 \% \\
            & UMAP  & LHRR  & LHRR     & 65.50 $\pm$ 0.07  \% & 81.49 $\pm$ 0.18 \% & 69.45 $\pm$ 0.22 \% \\
        \hline
    \end{tabular}
    }
    \label{tab:map_results_dogs_est_map}
\end{table*}

\begin{table*}[ht!]
    \centering
    \caption{\textbf{Precision at 100 (P@100)} results on the Dogs dataset across five independent executions varying the stochastic component of the framework (UMAP projection). The table reports the mean and standard deviation obtained for each configuration.}
    \resizebox{0.99\textwidth}{!}{
    \begin{tabular}{|c|lll|ccc|}
        \hline
        \multirow{2}{*}{\centering \textbf{Method}} & \multicolumn{3}{c}{\textbf{Method Specification}}   & \multicolumn{3}{|c|}{\textbf{Feature}} \\ \cline{2-5} \cline{6-7}
        & \textbf{Projection} & \textbf{Re-Rank} & \textbf{Post Re-Rank}   & \textbf{DinoV2} & \textbf{ResNet152} & \textbf{SwinTf} \\
        \hline
        \multirow{1}{*}{None - Baseline} 
        & -     & -     & -        & 63.57 \%  & 72.74 \%  & 59.83 \%  \\ 
        \hline
        \multirow{1}{*}{UMAP Only} 
        & UMAP  & -     & -       & 71.59 ± 0.08  \% & 83.67 ± 0.11  \% & 74.05 ± 0.08  \% \\ 
        \hline
        \multirow{3}{*}{Re-Rank Only} 
            & -     & CPRR  & -       & 71.17 \%  & \textbf{85.61 \%}  & 75.60 \%  \\
            & -     & RFE   & -       & 68.52 \%  & 81.24 \%  & 63.46 \%  \\
            & -     & LHRR  & -       & 70.53 \%  & 83.83 \%  & 69.40 \%  \\
        \hline
        \multirow{3}{*}{\shortstack{UMAP + Re-Rank \\\cite{kawai2024}}}
            & UMAP  & CPRR  & -        & 70.44 ± 0.12 \% & 82.55 ± 0.12 \% & 72.90 ± 0.14 \% \\
            & UMAP  & RFE   & -        & 70.63 ± 0.09 \% & 82.98 ± 0.15 \% & 73.62 ± 0.18 \% \\
            & UMAP  & LHRR  & -        & 70.41 ± 0.11 \% & 82.51 ± 0.14 \% & 72.99 ± 0.13 \% \\
        \hline
        \multirow{3}{*}{\shortstack{UMAP Rank Aggregation\\ (UMAP + Re-Rank)}}
            & UMAP  & CPRR  & -     &\textbf{72.77 $\pm$ 0.05 \%}  &85.27 $\pm$ 0.06  \% &75.58 $\pm$ 0.10  \% \\
            & UMAP  & RFE   & -     &71.98 $\pm$ 0.07  \% &84.33 $\pm$ 0.08  \% &71.73 $\pm$ 0.14  \% \\
            & UMAP  & LHRR  & -     &72.61 $\pm$ 0.04  \% &84.62 $\pm$ 0.08  \% &73.24 $\pm$ 0.10  \% \\
        \hline
        \multirow{9}{*}{\shortstack{UMAP Rank Aggregation\\ (UMAP + Re-Rank) \\ + Post Re-Rank}}
            & UMAP  & CPRR  & CPRR     &71.16 $\pm$ 0.06  \% &84.16 $\pm$ 0.09  \% &75.24 $\pm$ 0.12  \% \\
            & UMAP  & CPRR  & RFE      &71.01 $\pm$ 0.10  \% &83.38 $\pm$ 0.14  \% &72.20 $\pm$ 0.12  \% \\
            & UMAP  & CPRR  & LHRR     &71.41 $\pm$ 0.06  \% &84.36 $\pm$ 0.09  \% &74.99 $\pm$ 0.15  \% \\
            & UMAP  & RFE   & CPRR     &71.34 $\pm$ 0.15  \% &84.52 $\pm$ 0.13  \% &75.72 $\pm$ 0.20  \% \\
            & UMAP  & RFE   & RFE      &69.07 $\pm$ 0.17  \% &79.50 $\pm$ 0.11  \% &65.73 $\pm$ 0.26  \% \\
            & UMAP  & RFE   & LHRR     &71.03 $\pm$ 0.13  \% &84.07 $\pm$ 0.08  \% &73.68 $\pm$ 0.20  \%  \\
            & UMAP  & LHRR  & CPRR     &71.17 $\pm$ 0.08  \% &84.32 $\pm$ 0.11  \% &\textbf{75.80 $\pm$ 0.19}  \% \\
            & UMAP  & LHRR  & RFE      &70.30 $\pm$ 0.06  \% &82.10 $\pm$ 0.15  \% &69.07 $\pm$ 0.23  \% \\
            & UMAP  & LHRR  & LHRR     &71.30 $\pm$ 0.08  \% &84.49 $\pm$ 0.05  \% &75.16 $\pm$ 0.20  \% \\
        \hline
    \end{tabular}
    }
    \label{tab:map_results_dogs_est_p100}
\end{table*}

\end{document}